\documentclass[10pt,twocolumn,letterpaper]{article}

\usepackage[pagenumbers]{cvpr} %

\usepackage{cuted} %

\usepackage{currfile} %

\usepackage{caption} %
\usepackage{microtype}

\makeatletter
\renewcommand\paragraph{\@startsection{paragraph}{4}{\z@}{4pt \@plus 2pt \@minus 1pt}{-1em}{\normalfont\normalsize\bfseries}}
\makeatother

\usepackage[T1]{fontenc} %
\usepackage{listings}
\lstdefinestyle{harness}{
  language=Python,
  basicstyle=\ttfamily\scriptsize,
  keywordstyle=\color{cvprblue}\bfseries,
  commentstyle=\color{gray},
  stringstyle=\color{BrickRed},
  columns=fullflexible,
  keepspaces=true,
  showstringspaces=false,
  frame=tb,
  framerule=0.4pt,
  xleftmargin=2pt,
  aboveskip=4pt,
  belowskip=2pt,
  morekeywords={self},
}

\newcommand{\eo}{$^\dagger$}

\usepackage{tabularx}
\newcolumntype{Y}{>{\raggedright\arraybackslash}X}
\newcolumntype{L}[1]{>{\raggedright\arraybackslash}p{#1}}
\providecommand{\code}[1]{\texttt{#1}}
\usepackage{colortbl}
\definecolor{familyImageNet}{HTML}{4A78B5}
\definecolor{familyDINOv3}{HTML}{C9973F}
\definecolor{familyGFM}{HTML}{3E9C7A}
\newcommand{\modelcell}[2]{\cellcolor{family#1!18}#2}
\newcommand{\modelfamilylegend}{%
  \colorbox{familyImageNet!18}{\strut ImageNet}\quad
  \colorbox{familyDINOv3!18}{\strut DINOv3}\quad
  \colorbox{familyGFM!18}{\strut GFM}\quad\eo{}Pretrained on EO data%
}
\usepackage{tikz}
\usetikzlibrary{arrows.meta}
\newcommand{\computebar}[3]{%
  \begin{tikzpicture}[baseline=-0.3ex,x=5.5em,y=1ex]
    \path[use as bounding box] (0,0) rectangle (1,1);
    \fill[family#1] (0,0) rectangle (#2,0.9);
  \end{tikzpicture}%
  \hspace{0.4em}\makebox[2.4em][r]{#3}%
}

\definecolor{cvprblue}{rgb}{0.21,0.49,0.74}
\usepackage[pagebackref,breaklinks,colorlinks,allcolors=cvprblue]{hyperref}

\def\paperID{*****} %
\def\confName{CVPR}
\def\confYear{2027}

\title{geo-eval-harness: An Evaluation Harness for Geospatial Foundation Models}

\author{Nils Lehmann\textsuperscript{1}\thanks{Equal contribution.}, Caleb Robinson\textsuperscript{1}\footnotemark[1], Isaac Corley\textsuperscript{2}\footnotemark[1], Jennifer Marcus\textsuperscript{2},\\
Inbal Becker-Reshef\textsuperscript{1}, Juan M. Lavista Ferres\textsuperscript{1}, Hannah Kerner\textsuperscript{2,3}\\
\textsuperscript{1}Microsoft AI for Good Research Lab\\
\textsuperscript{2}Taylor Geospatial\quad \textsuperscript{3}Arizona State University
}

\begin{document}
\maketitle

\begin{figure*}[t]
	\centering
	\includegraphics[width=\textwidth]{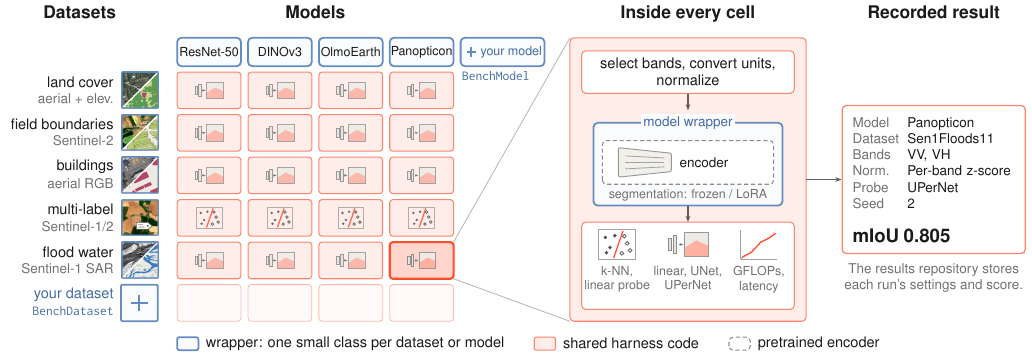}
	\caption{\textbf{\texttt{geo-eval-harness}} is an extensible \textit{evaluation harness} for earth observation (EO) models. We evaluate GFMs and general vision models (columns) on known EO datasets (rows) consisting of classification and dense prediction tasks. For a given run, the harness selects the bands that the model accepts, converts their units, and normalizes them in the way that the model expects. The harness fits a classification probe ($k$NN or linear) on frozen features, or trains a segmentation head (linear, UNet, or UPerNet) with a frozen or LoRA-adapted encoder. It also profiles inference GFLOPs and latency, and stores each run's versioned configuration and score in the results repository.}
	\label{fig:overview}
\end{figure*}

\begin{abstract}
Geospatial foundation model (GFM) research papers are plagued by inconsistent evaluation settings (e.g., different pre-processing and probes), making it difficult to separate performance from evaluation choices. We propose \texttt{geo-eval-harness}, an open-source framework for controlled model comparisons. We use this harness to evaluate 100 GFMs and general vision models on 6 classification and 14 segmentation datasets, using multiple probes on frozen or low-rank-adapted encoders.
We show that under controlled evaluation, general vision backbones are competitive on RGB classification and high-resolution segmentation, but GFMs shine on multispectral tasks. When we compare any two of the top-ten classification models on a single dataset, their 95\% confidence intervals overlap 73\% of the time. We show that bigger models are not always better: inference FLOPs correlate weakly with segmentation performance. 
Moreover, translating inference compute into large-area mapping workloads shows how small accuracy differences have outsized deployment cost differences: with frozen encoders on SpaceNet2, a 0.9-point mIoU difference corresponds to $5.9\times$ the compute and days of additional projected GPU time.
Our findings stress the importance of comparing models under explicit and fixed settings, and reporting uncertainty and deployment compute alongside accuracy. Our \texttt{geo-eval-harness} makes controlled experiments easy, fast, and fully reproducible. We release the code, configurations, and results at \url{https://github.com/torchgeo/geo-eval-harness}.
\end{abstract}

\section{Introduction}
Geospatial foundation models (GFMs) are developed to learn representations that transfer across Earth observation (EO) tasks, from agricultural monitoring and land-cover mapping to disaster response~\citep{lu2024vfmrs,jakubik2023prithvi}. These models differ in their pretraining methods and in their pretraining data, which varies in modality (optical, radar, hyperspectral), sensor (e.g., Sentinel-2 or aerial RGB), and spatial resolution (sub-meter to 30\,m)~\citep{lu2024vfmrs,rolf2024mission}. As the range of available models grows, assessing their usefulness in downstream tasks requires a practical way to compare new candidates under controlled conditions.

However, published results do not always provide a reliable starting point for such comparisons. In an audit of 152 GFM papers, \citet{corley2026noone} identified 47 cases in which the scores reported for the same model, benchmark, and protocol differed by at least five points across papers, including 18 that differed by at least ten points. The same audit found that 39\% of papers released no model weights and that papers making up 19\% of the corpus provided a public code repository but no released model~\citep{corley2026noone}. These findings mirror broader concerns about the reproducibility of empirical results in machine learning~\citep{pineau2021improving} and motivate developing a shared evaluation harness for GFMs. Language-model evaluation offers established precedents for this approach: HELM combines standardized evaluation conditions with public results and a reusable toolkit~\citep{liang2023holisticevaluationlanguagemodels}, while lm-evaluation-harness provides a unified framework for evaluating language models across diverse benchmarks with consistent metrics and protocols~\citep{lintang_sutawika_2026_22214970}. Marin extends this transparency to model development and publicly documents training and evaluation pipelines as well as logs for successful and failed experiments~\cite{hall2025marin}.

Within EO, GEO-Bench~\citep{lacoste2023geobench}, PANGAEA~\citep{marsocci2024pangaea}, and GEO-Bench-2~\citep{simumba2026geobench} have established common tasks and evaluation recipes for GFMs. However, building on these resources requires making their evaluation procedures easy to reuse as new models become available. Even when studies adopt the same datasets and protocols, evaluation pipelines that are built independently can differ enough to prevent direct comparison. One cause is the preprocessing needed to handle the sensor and spatial characteristics of EO data~\citep{rolf2024mission}. Freezing a model's weights still leaves choices about image resizing, normalization, spectral-band selection and ordering, feature pooling, optimization, and hyperparameter search, and these choices can change the measured quality of a representation. \citet{corley2024revisiting}, for example, report accuracy gains of 11 points on EuroSAT and 32 points on So2Sat from image resizing and normalization alone.

We present \texttt{geo-eval-harness}, an open-source harness for controlled comparisons between GFMs, to address this gap. Our contributions are:
\begin{enumerate}
    \item \textbf{A shared evaluation harness and comprehensive benchmark results for GFMs.} Most GFM papers evaluate with their own code, and the same model on the same dataset has been reported with scores ten or more points apart~\cite{corley2026noone}. Our shared harness is easily extensible -- a model or dataset is added by writing a small wrapper and we use it to evaluate 100 models on six classification and 14 segmentation datasets.
    \item \textbf{A study of evaluation sensitivity and uncertainty.} We show several common evaluation choices change benchmark scores. Switching from a $k$NN to a linear probe raises scores by 15.3 points on average and reorders the models (Kendall's $\tau=0.60$), and normalizing inputs with a model's pretraining statistics instead of the dataset's changes scores by up to 37.5 points. Any two of the ten best classification models also have overlapping 95\% confidence intervals on a single dataset 73\% of the time.
    \item \textbf{A controlled comparison of model accuracy and cost.} We find that inference compute is only weakly related to segmentation rank (Spearman $\rho=-0.11$), however this matters for downstream applications. On SpaceNet2, the best frozen model needs $5.9\times$ the compute of ConvNeXt-T DINOv3 to map 1{,}000{,}000\,km$^2$, for a 0.9-point gain in mIoU.
\end{enumerate}

\section{Related work}
\paragraph{EO benchmarks and evaluation software.}
GEO-Bench, PANGAEA, and GEO-Bench-2 establish shared datasets and protocols for evaluating transfer across EO tasks~\citep{lacoste2023geobench,marsocci2024pangaea,simumba2026geobench}. GEO-Bench-2 also supports band reordering and normalization, while TerraTorch provides configurable training, inference, and hyperparameter optimization for GFMs~\citep{gomes2025terratorch}. TorchGeo provides geospatial dataset interfaces and pretrained models~\citep{stewart2022torchgeo}. We build on these resources with a common experimental framework for studying representation transfer and evaluation sensitivity. To support reproducible comparisons as new checkpoints appear, models share the evaluation code while wrappers handle the various input requirements of a model explicit.

\paragraph{Efficient frozen-feature evaluation.}
PANGAEA supports linear and $k$NN classification probes, and GEO-Bench-2 supports frozen-backbone evaluation alongside its fine-tuning baseline. We focus on reducing the cost of repeated evaluations by caching embeddings and combining exact GPU nearest-neighbor search with full-batch L-BFGS logistic regression for classification problems. Reusing embeddings across these fits enables broad regularization searches without repeated encoder inference, while stored predictions support bootstrap interval estimation.

\paragraph{Evaluation sensitivity and uncertainty.}
Prior EO studies document sensitivity to preprocessing and discrepancies between published scores~\citep{corley2024revisiting,fuller2026badtables,corley2026noone}. Audits in metric learning and reinforcement learning similarly show how experimental procedures and uncertainty reporting affect conclusions about progress~\citep{musgrave2020metric,agarwal2021deep}. We measure these effects for GFMs with frozen weights (\cref{sec:choices}).

\paragraph{Deployment-scale compute comparisons.}
GEO-Bench-2 groups tasks by characteristics including resolution and spectral input, providing comparisons relevant to different applications~\citep{simumba2026geobench}. We extend this perspective by relating frozen and LoRA-adapted segmentation accuracy to the resources required for large-area mapping. Alongside inference profiles, we use input-specific FLOPs and ground footprints to estimate the compute needed to cover a physical area.

\section{\texttt{geo-eval-harness}}
\label{sec:harness}

\texttt{geo-eval-harness} is an open-source harness that evaluates pretrained models on EO benchmark datasets with one shared pipeline (\cref{fig:overview}). Each model and each dataset enters through a small wrapper (supplementary \cref{lst:wrappers}), and the harness handles band selection, normalization, feature extraction, probing, cost measurement, and result logging. Every choice in this pipeline is a configuration value. Supplementary \cref{tab:settings} lists these settings, which we use in \cref{sec:choices} to measure how much they influence ranking differences. The following subsections describe the wrappers (\cref{sec:wrappers}), the probes and segmentation training (\cref{sec:probes}), the cost measurement (\cref{sec:cost}), and the datasets and models (\cref{sec:suite}).

\subsection{Model and dataset wrappers}
\label{sec:wrappers}

Any pretrained model expects its inputs to come from the same distribution that was used during pretraining, i.e. with a fixed set of bands in a fixed order and values in a particular range\footnote{Several GFMs differ in their requirements for spectral bands, channel ordering, pixel-value scaling, normalization, and spatial resolution.}. ImageNet models, for example, expect red, green, and blue values scaled to $[0,1]$ and standardized with the ImageNet channel means and standard deviations~\cite{deng2009imagenet}. Prithvi~\cite{jakubik2023prithvi} instead expects six Harmonized Landsat Sentinel-2 bands (blue, green, red, narrow near infrared, and two shortwave infrared bands) as reflectance multiplied by 10,000 and standardized with its own pretraining statistics.

When creating benchmark datasets, authors must make choices on how to store and present EO data in an ``ML friendly'' format. For example, the Sentinel-2 Level-1C and Level-2A products store reflectance values multiplied by 10,000 as 16-bit unsigned integers, but neural networks require floating point numbers that are typically centered at 0.  GEO-Bench EuroSAT~\cite{lacoste2023geobench,helber2019eurosat} keeps the Sentinel-2 integers unchanged, So2Sat~\cite{zhu2020so2sat} divides them by 10,000, and the GEO-Bench-2 version of HLS Burn Scars~\cite{jakubik2023prithvi,simumba2026geobench} stores reflectance between 0 and 1. Sen1Floods11~\cite{bonafilia2020sen1floods11} stores Sentinel-1 backscatter in decibels, while Kuro Siwo~\cite{bountos2024kurosiwo} stores it as linear power. Dataset releases do not always state the units of the stored values, and model releases do not always state the units and band order that their normalization statistics assume (supplementary \cref{fig:normalization}).
Similarly, the bands a model sees, and their order, also matter. A network applies its first-layer weights by channel position, so a model trained on red, green, and blue that receives blue, green, and red still runs but treats blue as red. Datasets store bands in different orders. GEO-Bench EuroSAT and PV4GER store blue before red, while Fields of The World (FotW)~\cite{kerner2025fotw} stores red first.

An evaluation harness for EO data therefore has to give every model each dataset in the form that the model expects, which \texttt{geo-eval-harness} does in three steps (\cref{fig:overview}).
\begin{enumerate*}[label=(\arabic*)]
	\item Each dataset declares its task, classes, split sizes, and one \texttt{BandSpec} object per channel, with the sensor, a canonical band name, the center wavelength, and statistics measured on its training split.
	\item For a given model, the harness selects the bands it requires by name and builds a new model wrapper for each dataset from the \texttt{BandSpec} of the selected bands. The wrapper places the bands in the order the model checkpoint expects, so it can read their wavelengths, as DOFA~\cite{xiong2024dofa} does, or route each sensor to its own encoder. RGB checkpoints that receive more bands use an input layer whose extra channels reuse the RGB weights in the way described in the timm library~\cite{wightman2019timm}. If the configuration isn't valid, then the harness will reject that run.
	\item Finally, the harness normalizes the dataset values (supplementary \cref{sec:supp-normalization}). By default it standardizes each band with the dataset's training statistics. For models that we evaluate with their pretraining normalization (supplementary \cref{sec:supp-rules}), it instead converts the dataset values to the units used in pretraining (e.g. from Sentinel-2 integers to reflectance), and then applies the published pretraining statistics.
\end{enumerate*}

\subsection{Evaluating models}
\label{sec:probes}

\paragraph{Classification probing.}
To evaluate models on classification tasks we use $k$NN and linear probes on their frozen dataset embeddings.  We implement both probes on the GPU, so that a full regularization search over a dataset's embeddings takes seconds, and we match scikit-learn's~\cite{pedregosa2011scikit} model definition, so that our scores can be compared with probes fit in scikit-learn.

Specifically, our $k$NN probe searches the training embeddings exactly with FAISS~\cite{johnson2019faiss} under Euclidean distance and takes an unweighted vote of the $k=5$ nearest neighbors. In multi-label datasets each label is predicted when more than half of the neighbors carry it.
Our linear probe minimizes scikit-learn's logistic-regression objective: the mean cross-entropy plus $\|W\|^2/(2Cn)$ for $n$ training samples without a penalty on the bias. For multi-label data  we use a sigmoid cross-entropy for each label and scale the penalty so that the solution equals fitting one binary scikit-learn model per label. We minimize the objective in PyTorch with full-batch L-BFGS~\cite{liu1989lbfgs} and a strong Wolfe line search on the GPU, starting from zero weights. In our tests, its predicted probabilities agree with scikit-learn's to within $10^{-3}$ for $C$ from $10^{-2}$ to $10^{2}$, for single-label and multi-label data.

By default, we resize images to $224\times224$ pixels with bilinear interpolation and extract one embedding per image, averaging over tokens for transformer outputs. Model-specific resizing and pooling exceptions are listed in supplementary \cref{tab:app-models}. We extract the embeddings of each split once and reuse them for both probes and the regularization search. We fit one linear probe for each of 40 values of $C$ spaced logarithmically from $10^{-6}$ to $10^{4}$, choose $C$ by the validation score, and refit on the union of the training and validation splits. L-BFGS requires no learning-rate search, and we only tune $C$ while keeping the remaining solver settings fixed. We report top-1 accuracy for single-label datasets and micro mAP for multi-label datasets. Lastly, we compute a 95\% interval for each score from 200 bootstrap resamples of the test predictions.

\paragraph{Segmentation probing.}
For segmentation tasks, we train a UPerNet decoder~\cite{xiao2018upernet} on four backbone feature maps. Hierarchical encoders provide this pyramid directly, and for plain ViTs a trainable neck rescales the equal-resolution maps to strides 32, 16, 8, and 4, following the simple feature pyramid of Li \etal~\cite{li2022vitdet}. We evaluate three regimes. With a frozen encoder, we train the decoder and neck only, caching backbone features when memory permits. With LoRA~\cite{hu2022lora}, we use PEFT~\cite{peft} to insert rank-8 adapters into the attention and MLP projections and train them jointly with the decoder. With full fine-tuning, which we report only in \cref{sec:choices}, we update all encoder weights at a learning rate of $10^{-5}$. We resize inputs to $224\times224$ pixels on all datasets except four whose fine-grained labels would lose detail at that size. On these we use 256 pixels for FotW and 512 pixels for LoveDA, SpaceNet2, and SpaceNet7. For the main segmentation comparison, we train with AdamW~\cite{loshchilov2019adamw}, no weight decay, a batch size of 16, and a decoder learning rate of $10^{-3}$, with cosine decay over 30 epochs. LoRA adapters use a learning rate of $10^{-4}$ or $5\times10^{-4}$, depending on the dataset (supplementary \cref{sec:supp-rules}). We restore the checkpoint with the highest validation mean intersection-over-union (mIoU) and report the test set mean over three seeds. To examine decoder sensitivity, we also compare UPerNet with UNet~\cite{ronneberger2015unet} and with per-location linear, FPN~\cite{lin2017fpn}, and DPT~\cite{ranftl2021dpt} heads.

\subsection{Computational cost}
\label{sec:cost}

In the general computer vision domain, controlled experiments often allow for fair comparisons between models by standardizing training data and architecture. In EO, however, GFMs often differ in their pretraining data, training duration, and input processing alongside the learning method itself. As pointed out by \citet{fuller2026badtables}, even models described as ViT-Base can use different image sizes, patch sizes, and spectral channel groupings, making their shared architecture label an insufficient basis for comparison.

For widely used ViT architectures among GFMs, these choices affect the number of tokens a model processes and therefore its computational cost, often without a comparable change in parameter count. We therefore report floating-point operations (FLOPs) for explicit input configurations alongside downstream performance, so that gains associated with more expensive processing are visible in the comparison.

PyTorch's operator-level counter works from tensor shapes, so we count FLOPs on one forward pass over a synthetic input. We measure each model at 224, 256, and 512 pixels with RGB and 12-band Sentinel-2 inputs, once for the backbone alone and once with the segmentation neck and decoder. We count each multiply--add as two operations, so our values are about twice those of tools that report multiply--accumulates. We also record parameter count, throughput, median latency, and peak GPU memory in float32 on one NVIDIA RTX PRO 6000 Blackwell (see supplementary \cref{app:compute}).

The compute bars in \cref{tab:seg-lora} and the sensor-group comparison in supplementary \cref{fig:sensor-accuracy} use standardized $224\times224$ RGB reference inputs. In contrast, the dataset-specific mapping projections use the evaluation input size and the ground footprint of each dataset tile. Resizing a tile changes its effective GSD but preserves its ground footprint.

\subsection{Datasets and models}
\label{sec:suite}

\begin{table*}[t]
	\centering
	\setlength{\tabcolsep}{2pt}
	\resizebox{\textwidth}{!}{%
\begin{tabular}{@{}lrrrrrrrrrrrrrrrrrr@{}}
\toprule
 & \multicolumn{1}{c}{HLS} & \multicolumn{3}{c}{Sentinel-2} & \multicolumn{1}{c}{PlanetScope} & \multicolumn{4}{c}{Aerial / VHR} & \multicolumn{2}{c}{Sentinel-1} & \multicolumn{2}{c}{VHR SAR} & \multicolumn{1}{c}{Mixed SAR} & \multicolumn{3}{c}{Summary} & \multicolumn{1}{c}{Inference} \\
\cmidrule(lr){2-2}\cmidrule(lr){3-5}\cmidrule(lr){6-6}\cmidrule(lr){7-10}\cmidrule(lr){11-12}\cmidrule(lr){13-14}\cmidrule(lr){15-15}\cmidrule(lr){16-18}\cmidrule(l){19-19}
Model & BurnScars & CloudSEN12 & FotW & \shortstack{S1Floods\\S2} & SpaceNet7 & FLAIR\#2 & HOTOSM & LoveDA & SpaceNet2 & KuroSiwo & \shortstack{S1Floods\\S1} & \shortstack{BRIGHT\\3ch} & \shortstack{OEM\\SAR} & CaFFe & Mean & Rank & Frozen & GFLOPs / image \\
\midrule
\modelcell{DINOv3}{DINOv3 ViT-L/16} & 89.7 & 72.2 & 60.8 & 76.3 & 58.6 & 56.3 & \textbf{67.4} & \textbf{52.7} & 88.5 & 58.1 & 79.7 & 38.9 & 33.7 & \textbf{70.3} & 64.5 & 6.0 & 7.4 & \computebar{DINOv3}{0.551965}{157.9} \\
\modelcell{DINOv3}{DINOv3-SAT ViT-L/16\eo} & 89.3 & 71.9 & 60.4 & 74.8 & 58.9 & \textbf{56.3} & 65.8 & 51.2 & 87.9 & \textbf{58.4} & 79.7 & \textbf{41.3} & \textbf{34.5} & 68.7 & 64.2 & 6.9 & 9.0 & \computebar{DINOv3}{0.551965}{157.9} \\
\modelcell{ImageNet}{SwinV2-L (22k)} & 88.8 & 71.8 & 60.4 & 77.9 & 59.0 & 53.4 & 67.2 & 49.6 & 87.7 & 58.0 & \textbf{81.2} & 39.4 & 32.1 & 65.4 & 63.7 & 7.6 & 10.3 & \computebar{ImageNet}{0.332060}{95.0} \\
\modelcell{DINOv3}{ConvNeXt-L DINOv3} & 88.2 & 71.8 & 60.3 & 74.7 & 59.8 & 55.2 & 67.1 & 52.6 & \textbf{88.7} & 55.6 & 79.4 & 38.9 & 32.4 & 66.7 & 63.7 & 9.1 & 11.4 & \computebar{DINOv3}{0.313086}{89.5} \\
\modelcell{ImageNet}{SwinV2-B (22k)} & 89.3 & 71.9 & 59.7 & 77.2 & 58.9 & 52.3 & 66.6 & 50.8 & 87.6 & 57.4 & 80.5 & 38.3 & 31.0 & 63.0 & 63.2 & 9.6 & 11.5 & \computebar{ImageNet}{0.190898}{54.6} \\
\modelcell{DINOv3}{ConvNeXt-B DINOv3} & 86.9 & 71.1 & 59.7 & 72.8 & 59.7 & 54.4 & 66.8 & 52.5 & 88.4 & 57.8 & 78.9 & 38.7 & 32.4 & 64.8 & 63.2 & 10.6 & 15.0 & \computebar{DINOv3}{0.179087}{51.2} \\
\modelcell{DINOv3}{ConvNeXt-S DINOv3} & 86.5 & 70.8 & 59.4 & 72.2 & 59.4 & 53.8 & 65.9 & 51.3 & 88.3 & 58.3 & 79.1 & 39.1 & 32.9 & 64.7 & 63.0 & 11.9 & 14.3 & \computebar{DINOv3}{0.131896}{37.7} \\
\modelcell{ImageNet}{Swin-L} & 88.7 & 70.9 & 59.6 & 76.6 & 59.9 & 51.9 & 65.7 & 50.2 & 87.3 & 57.3 & 80.5 & 37.6 & 29.9 & 64.0 & 62.9 & 12.1 & 12.4 & \computebar{ImageNet}{0.313885}{89.8} \\
\modelcell{GFM}{DEO Swin} & 87.5 & 71.2 & 59.3 & 87.9 & 60.3 & 53.3 & 65.5 & 48.1 & 87.2 & 56.9 & 79.2 & 40.0 & 31.2 & 62.5 & 63.6 & 12.9 & 7.4 & \computebar{GFM}{0.250002}{71.5} \\
\modelcell{GFM}{OlmoEarth v1.2 base} & 92.4 & \textbf{72.7} & \textbf{63.9} & \textbf{88.8} & \textbf{61.2} & 45.7 & 63.0 & 43.4 & 86.0 & 57.8 & 81.1 & 36.3 & 27.0 & 54.2 & 62.4 & 14.3 & 17.7 & \computebar{GFM}{0.943396}{269.8} \\
\modelcell{ImageNet}{SwinV2-S} & 88.7 & 70.9 & 58.9 & 73.8 & 58.2 & 51.0 & 66.5 & 48.7 & 87.1 & 57.0 & 79.6 & 39.2 & 31.3 & 62.7 & 62.4 & 14.7 & 17.0 & \computebar{ImageNet}{0.138429}{39.6} \\
\modelcell{GFM}{TerraMind v1 large} & 90.8 & 72.1 & 60.2 & 87.0 & 57.1 & 47.0 & 62.7 & 45.6 & 86.2 & 57.7 & 80.4 & 37.0 & 31.1 & 61.3 & 62.6 & 15.3 & 19.9 & \computebar{GFM}{0.540655}{154.6} \\
\modelcell{DINOv3}{ConvNeXt-T DINOv3} & 86.7 & 70.2 & 58.3 & 72.4 & 60.0 & 51.3 & 65.9 & 51.7 & 88.0 & 57.3 & 78.6 & 37.7 & 31.3 & 61.4 & 62.2 & 15.9 & 17.4 & \computebar{DINOv3}{0.102326}{29.3} \\
\modelcell{ImageNet}{ConvNeXt-L} & 87.0 & 70.4 & 59.9 & 74.3 & 59.1 & 50.0 & 66.1 & 49.4 & 87.7 & 57.0 & 78.4 & 37.5 & 30.8 & 62.5 & 62.2 & 16.1 & 17.9 & \computebar{ImageNet}{0.313086}{89.5} \\
\modelcell{ImageNet}{SwinV2-T} & 87.4 & 70.4 & 58.7 & 76.3 & 59.2 & 50.5 & 65.1 & 49.7 & 86.8 & 56.9 & 79.7 & 38.6 & 30.4 & 64.1 & 62.4 & 16.1 & 14.1 & \computebar{ImageNet}{0.105162}{30.1} \\
\modelcell{ImageNet}{Swin-B} & 87.9 & 70.8 & 59.0 & 76.0 & 59.4 & 51.0 & 66.2 & 48.9 & 87.0 & 56.3 & 79.3 & 37.2 & 30.3 & 62.7 & 62.3 & 16.4 & 15.5 & \computebar{ImageNet}{0.179620}{51.4} \\
\modelcell{GFM}{OlmoEarth v1.2 small} & 92.1 & 72.2 & 62.9 & 88.6 & 60.2 & 44.1 & 61.9 & 43.1 & 85.6 & 57.1 & 79.7 & 35.7 & 27.9 & 52.9 & 61.7 & 17.1 & 20.7 & \computebar{GFM}{0.475114}{135.9} \\
\modelcell{ImageNet}{ConvNeXt-S} & 87.4 & 70.6 & 59.3 & 73.5 & 58.8 & 49.8 & 65.5 & 49.4 & 87.8 & 56.5 & 79.3 & 38.0 & 29.2 & 61.1 & 61.9 & 18.2 & 24.8 & \computebar{ImageNet}{0.131896}{37.7} \\
\modelcell{ImageNet}{ConvNeXt-B} & 87.4 & 70.3 & 59.2 & 73.7 & 59.6 & 49.2 & 65.5 & 49.8 & 87.5 & 56.6 & 78.8 & 36.8 & 30.1 & 60.8 & 61.8 & 19.4 & 22.5 & \computebar{ImageNet}{0.179087}{51.2} \\
\modelcell{GFM}{OlmoEarth v1.3 base} & \textbf{92.8} & 71.6 & 62.0 & 87.9 & 59.3 & 43.3 & 62.6 & 39.9 & 84.9 & 54.5 & 80.3 & 36.0 & 25.5 & 51.0 & 60.8 & 19.8 & 19.9 & \computebar{GFM}{0.943396}{269.8} \\
\modelcell{ImageNet}{Swin-S} & 85.8 & 70.2 & 58.6 & 74.4 & 58.1 & 50.6 & 65.4 & 48.8 & 87.1 & 56.9 & 79.3 & 36.9 & 30.7 & 61.7 & 61.8 & 20.1 & 20.9 & \computebar{ImageNet}{0.132296}{37.8} \\
\modelcell{GFM}{Panopticon} & 89.7 & 70.2 & 59.0 & 86.6 & 57.6 & 49.4 & 61.5 & 45.8 & 85.0 & 56.9 & 79.6 & 37.9 & 29.6 & 54.4 & 61.7 & 20.6 & 18.7 & \computebar{GFM}{0.348252}{99.6} \\
\modelcell{ImageNet}{Swin-T} & 86.7 & 69.5 & 58.4 & 74.0 & 58.4 & 48.3 & 65.8 & 49.7 & 86.6 & 56.3 & 79.2 & 37.3 & 30.1 & 62.7 & 61.7 & 20.6 & 18.2 & \computebar{ImageNet}{0.102571}{29.3} \\
\modelcell{ImageNet}{ConvNeXt-T} & 87.4 & 70.1 & 58.6 & 73.1 & 59.3 & 49.2 & 64.4 & 49.5 & 87.5 & 57.1 & 78.0 & 35.6 & 30.4 & 61.6 & 61.6 & 20.9 & 25.0 & \computebar{ImageNet}{0.102326}{29.3} \\
\modelcell{GFM}{DOFA large} & 90.4 & 68.7 & 59.2 & 84.7 & 56.3 & 50.4 & 65.2 & 48.1 & 86.1 & 54.9 & 76.9 & 37.0 & 30.5 & 61.7 & 62.1 & 21.0 & 22.4 & \computebar{GFM}{0.544251}{155.6} \\
\modelcell{GFM}{OlmoEarth v1.2 tiny} & 92.2 & 70.9 & 61.2 & 88.5 & 57.9 & 42.7 & 61.2 & 39.5 & 84.2 & 56.4 & 80.1 & 35.1 & 25.0 & 53.6 & 60.6 & 21.1 & 22.1 & \computebar{GFM}{0.342842}{98.0} \\
\modelcell{ImageNet}{ViT-L/16} & 87.9 & 69.6 & 59.0 & 72.0 & 56.1 & 51.1 & 65.6 & 47.7 & 86.3 & 57.0 & 78.3 & 35.5 & 31.3 & 61.9 & 61.4 & 21.3 & 19.6 & \computebar{ImageNet}{0.542970}{155.3} \\
\modelcell{GFM}{TerraMind v1 base} & 90.8 & 71.7 & 58.5 & 86.9 & 57.7 & 44.1 & 62.5 & 44.7 & 85.8 & 53.8 & 79.2 & 36.7 & 30.6 & 56.8 & 61.4 & 21.5 & 19.9 & \computebar{GFM}{0.217474}{62.2} \\
\modelcell{ImageNet}{ViT-B/16} & 86.3 & 69.9 & 58.6 & 72.2 & 56.7 & 50.5 & 64.2 & 48.9 & 86.1 & 57.3 & 78.9 & 36.4 & 30.4 & 60.3 & 61.2 & 22.7 & 22.2 & \computebar{ImageNet}{0.218119}{62.4} \\
\modelcell{GFM}{DOFA base} & 90.5 & 68.0 & 57.7 & 85.0 & 56.0 & 48.7 & 64.9 & 48.7 & 85.9 & 55.7 & 78.1 & 36.4 & 30.8 & 59.4 & 61.8 & 23.0 & 24.5 & \computebar{GFM}{0.219224}{62.7} \\
\modelcell{GFM}{OlmoEarth v1.2 nano} & 92.4 & 69.5 & 59.3 & 87.9 & 57.4 & 39.0 & 59.8 & 35.9 & 83.0 & 54.2 & 79.2 & 33.5 & 25.5 & 52.8 & 59.2 & 26.1 & 24.8 & \computebar{GFM}{0.293862}{84.0} \\
\modelcell{ImageNet}{ViT-S/16} & 86.1 & 68.5 & 56.8 & 70.8 & 55.9 & 48.2 & 63.8 & 47.1 & 85.2 & 57.3 & 78.1 & 34.6 & 28.2 & 59.2 & 60.0 & 27.7 & 29.3 & \computebar{ImageNet}{0.109731}{31.4} \\
\modelcell{GFM}{SMARTIES large} & 89.3 & 68.0 & 55.6 & 83.8 & 56.0 & 42.0 & 62.1 & 40.2 & 83.7 & 56.0 & 77.6 & 34.4 & 22.1 & 52.9 & 58.8 & 28.6 & 28.1 & \computebar{GFM}{0.542970}{155.3} \\
\modelcell{GFM}{SMARTIES base} & 88.8 & 67.3 & 54.9 & 85.0 & 56.2 & 39.9 & 61.8 & 36.9 & 83.3 & 54.6 & 77.6 & 33.7 & 20.8 & 52.2 & 58.1 & 29.8 & 29.0 & \computebar{GFM}{0.218119}{62.4} \\
\bottomrule
\end{tabular}
}
	\par\smallskip
	{\scriptsize\modelfamilylegend}
	\caption{\textbf{Semantic segmentation with LoRA.} mIoU ($\times100$) averaged over three seeds on the 14 core datasets. Rank is the average per-dataset rank among the 34 models with LoRA results on all datasets, and Frozen is the same model's average rank with a frozen encoder among 37 models. Bars show backbone and UPerNet GFLOPs for a reference $224\times224$ RGB input; evaluation input sizes and band counts can differ.. Bold marks the best score per dataset. VHR is very-high resolution.}
	\label{tab:seg-lora}
\end{table*}

\paragraph{Datasets.}
In early experiments, most models scored above 95\% on GEO-Bench's PV4GER and Brick Kiln, with a median linear-probe accuracy of about 96\% on each. We leave both out of the reported results because scores this close do little to separate models, although we do not take them to mean that the underlying tasks are solved. Our reported results use 6 classification and 14 segmentation datasets, which we call the \emph{core datasets} (supplementary \cref{tab:supp-datasets}). Supplementary \cref{app:datasets} notes where our splits, labels, and bands differ from the original releases.

\paragraph{Models.}
We evaluate 100 models with complete RGB classification results: 45 pretrained on EO data, 53 pretrained on natural images, and two baselines without pretraining, random convolutional features (RCF)~\cite{rolf2021mosaiks} and per-band image statistics. The EO models include OlmoEarth~\cite{herzog2025olmoearth} v1 to v1.3 from nano to large, TerraMind~\cite{jakubik2025terramind}, Prithvi~\cite{jakubik2023prithvi,szwarcman2024prithvi2}, Clay~\cite{clay2024}, DOFA~\cite{xiong2024dofa}, Panopticon~\cite{waldmann2025panopticon}, DEO~\cite{wolf2026deo}, ScaleMAE~\cite{reed2023scalemae}, SatlasNet~\cite{bastani2023satlaspretrain}, SeCo~\cite{manas2021seco}, GASSL~\cite{ayush2021gassl}, SSL4EO MoCo~\cite{wang2023ssl4eo}, SMARTIES~\cite{sumbul2025smarties}, UniverSat~\cite{perron2026universat}, and DINOv3 trained on satellite images~\cite{simeoni2025dinov3}. CROMA~\cite{fuller2023croma} and TerraFM~\cite{danish2025terrafm} have results only with all bands, so they are absent from the RGB comparisons. The general vision models are CNNs and transformers~\cite{he2016resnet,huang2017densenet,tan2019efficientnet,tan2021efficientnetv2,howard2019mobilenetv3,radosavovic2020regnet,simonyan2015vgg,liu2022convnext,woo2023convnextv2,liu2021swin,liu2022swinv2,dosovitskiy2021vit,touvron2021deit,tu2022maxvit} from the timm library~\cite{wightman2019timm}, and the web-image DINOv3 ViT-L and ConvNeXt models~\cite{simeoni2025dinov3}. Supplementary \cref{tab:supp-models} lists every model with its parameter count, GFLOPs, and throughput.

\section{Results and analysis}
\label{sec:analysis}
\label{sec:results}

We first compare models on segmentation and classification (\cref{sec:analysis:settings}), then relate segmentation accuracy to the compute needed to map large areas (\cref{sec:mapping}), and finally measure how single evaluation choices change scores and rankings (\cref{sec:choices}). We focus on segmentation because its dense predictions correspond more directly to mapping applications such as flood extent, field boundaries, and building footprints. We benchmark classification because it is widely used in GFM evaluations and checks whether representations that transfer well to dense prediction also support image-level tasks.

\subsection{Model performance across evaluation settings}
\label{sec:analysis:settings}

We group models into ImageNet-pretrained checkpoints, DINOv3 checkpoints, and GFMs. The satellite-trained DINOv3 (DINOv3-SAT) is grouped with the DINOv3 checkpoints and marked with \eo{} because it was pretrained on EO data. Because an aggregate ranking depends on which tasks are included and how scores are combined~\cite{dehghani2021benchmarklottery,zhang2024inherent,hardt2026emerging}, we report per-dataset scores grouped by sensor, with separate RGB and all-band classification comparisons. On multispectral segmentation datasets, general vision models use RGB inputs, while multispectral GFMs receive all bands they support.

\paragraph{Segmentation.}
Among the 34 models with LoRA results on all 14 core segmentation datasets (\cref{tab:seg-lora}), DINOv3 ViT-L/16 has the best average rank (6.0), followed by DINOv3-SAT ViT-L/16 (6.9) and the ImageNet SwinV2-L (22k) (7.6). DEO Swin (ninth) and OlmoEarth v1.2 base (tenth) are the highest-ranked GFMs, so the ten best models comprise five DINOv3, three ImageNet, and two GFM checkpoints. Inference cost is only weakly related to this order, with a Spearman correlation of $-0.11$ between rank and total cost, which spans 29 to 270 GFLOPs per $224\times224$ image. ConvNeXt-S DINOv3, for instance, ranks above OlmoEarth v1.2 base (11.9 against 14.3) while requiring $7.2\times$ fewer GFLOPs.

\paragraph{Classification.}
\begin{table}[t]
	\centering
	\setlength{\tabcolsep}{2pt}
	\resizebox{\linewidth}{!}{%
\begin{tabular}{@{}lrrrrrrrrr@{}}
\toprule
 & \multicolumn{1}{c}{Landsat} & \multicolumn{3}{c}{Sentinel-2} & \multicolumn{2}{c}{Aerial / VHR} & \multicolumn{3}{c}{Summary} \\
\cmidrule(lr){2-2}\cmidrule(lr){3-5}\cmidrule(lr){6-7}\cmidrule(l){8-10}
Model & ForestNet & BENv2 & So2Sat & EuroSAT-Sp & TreeSatAI & RESISC45 & Mean & Rank & $k$NN \\
\midrule
\modelcell{GFM}{DOFA large} & 57.5 & 81.3 & 67.7 & \textbf{96.5} & 62.3 & 93.8 & 76.5 & 3.8 & 9.5 \\
\modelcell{DINOv3}{DINOv3 ViT-L/16} & 55.6 & 81.6 & 67.2 & 95.4 & \textbf{65.2} & \textbf{95.7} & 76.8 & 4.4 & 11.8 \\
\modelcell{GFM}{DEO Swin (RGB)} & 56.1 & \textbf{83.7} & 67.2 & 95.0 & 64.7 & 93.1 & 76.6 & 4.9 & 3.7 \\
\modelcell{DINOv3}{DINOv3-SAT ViT-L/16\eo} & \textbf{57.6} & 83.4 & 67.2 & 95.0 & 65.1 & 91.7 & 76.7 & 5.2 & 6.8 \\
\modelcell{ImageNet}{ViT-L/16} & 56.0 & 80.6 & 66.2 & 95.0 & 62.6 & 93.1 & 75.6 & 7.9 & 12.6 \\
\modelcell{GFM}{DOFA base} & 55.6 & 80.1 & 65.1 & 95.2 & 61.2 & 91.6 & 74.8 & 10.1 & 16.8 \\
\modelcell{GFM}{Panopticon} & 55.0 & 82.9 & \textbf{69.2} & 94.7 & 61.0 & 89.2 & 75.3 & 13.1 & 5.9 \\
\modelcell{ImageNet}{Swin-L} & 56.3 & 78.2 & 64.0 & 94.4 & 62.0 & 92.9 & 74.6 & 15.7 & 38.1 \\
\modelcell{ImageNet}{SwinV2-L (22k)} & 53.2 & 79.6 & 62.6 & 95.1 & 61.8 & 92.3 & 74.1 & 15.8 & 32.8 \\
\modelcell{ImageNet}{ConvNeXt-L} & 51.8 & 79.3 & 63.1 & 95.1 & 62.9 & 93.5 & 74.3 & 16.7 & 28.5 \\
\modelcell{ImageNet}{ConvNeXt-B} & 53.7 & 78.4 & 62.4 & 95.2 & 60.2 & 92.0 & 73.6 & 18.5 & 24.8 \\
\modelcell{GFM}{TerraMind v1 base} & 56.6 & 82.3 & 65.8 & 93.9 & 59.3 & 86.4 & 74.1 & 22.8 & 28.1 \\
\modelcell{ImageNet}{EfficientNet-B4} & 52.6 & 77.7 & 64.6 & 95.3 & 59.8 & 89.5 & 73.2 & 23.0 & 37.6 \\
\modelcell{GFM}{TerraMind v1 large} & 56.4 & 83.3 & 62.4 & 93.9 & 59.4 & 87.0 & 73.7 & 23.8 & 35.9 \\
\modelcell{GFM}{Clay v1.5 large} & 54.2 & 81.6 & 64.5 & 93.2 & 63.0 & 87.3 & 74.0 & 25.1 & 53.9 \\
\modelcell{ImageNet}{SwinV2-B (22k)} & 53.2 & 77.9 & 61.2 & 94.8 & 60.3 & 90.8 & 73.0 & 25.2 & 43.5 \\
\modelcell{ImageNet}{DeiT-B/16} & 52.0 & 78.9 & 61.5 & 94.6 & 59.7 & 90.6 & 72.9 & 27.1 & 19.7 \\
\modelcell{ImageNet}{ConvNeXtV2-B} & 50.5 & 78.6 & 62.0 & 94.9 & 59.5 & 90.8 & 72.7 & 27.8 & 37.8 \\
\modelcell{ImageNet}{Swin-B} & 52.5 & 77.2 & 61.0 & 94.8 & 60.2 & 91.0 & 72.8 & 28.1 & 47.6 \\
\modelcell{DINOv3}{ConvNeXt-L DINOv3} & 50.9 & 79.9 & 60.1 & 93.8 & 60.8 & 92.9 & 73.1 & 28.7 & 34.1 \\
\modelcell{GFM}{MoCo ResNet-50 (RGB)} & 53.5 & 81.8 & 59.3 & 95.9 & 57.8 & 87.4 & 72.6 & 28.8 & 19.8 \\
\modelcell{DINOv3}{ConvNeXt-B DINOv3} & 50.2 & 78.6 & 61.5 & 94.8 & 59.4 & 91.9 & 72.7 & 28.9 & 44.6 \\
\modelcell{DINOv3}{ConvNeXt-T DINOv3} & 54.5 & 77.2 & 62.0 & 94.7 & 58.1 & 89.5 & 72.7 & 29.2 & 53.1 \\
\modelcell{GFM}{ScaleMAE large} & 54.5 & 78.4 & 55.9 & 94.3 & 60.6 & 92.5 & 72.7 & 30.2 & 33.2 \\
\modelcell{ImageNet}{ConvNeXt-T} & 54.4 & 76.8 & 62.8 & 94.6 & 58.6 & 89.3 & 72.7 & 30.3 & 51.0 \\
\midrule
\modelcell{ImageNet}{ResNet-50} & 52.1 & 77.2 & 56.6 & 94.2 & 57.8 & 88.8 & 71.1 & 46.5 & 33.5 \\
Random conv. features & 46.9 & 68.2 & 33.7 & 65.9 & 48.5 & 57.1 & 53.4 & 96.5 & 99.0 \\
Image statistics & 34.8 & 47.6 & 35.6 & 67.4 & 36.5 & 35.9 & 43.0 & 99.7 & 98.0 \\
\bottomrule
\end{tabular}}
	\par\smallskip
	{\scriptsize\modelfamilylegend\quad Unshaded: no pretraining.}
	\caption{\textbf{RGB classification with a linear probe.} Top-1 accuracy, or micro mAP for BENv2 and TreeSatAI ($\times100$), on the six core datasets for the 25 best models and three references. Rank and $k$NN are the average per-dataset ranks with a linear and a $k$NN probe, respectively, among the same 100 models. Bold marks the best score per dataset.}
	\label{tab:cls}
\end{table}

DOFA large and DINOv3 ViT-L/16 lead the RGB linear-probe comparison with average ranks of 3.8 and 4.4 over the six core datasets (\cref{tab:cls}). The ten best models comprise four GFMs (DOFA large, DOFA base, DEO Swin, and Panopticon), the two DINOv3 ViT-L/16 models, and four ImageNet models (ViT-L/16, Swin-L, SwinV2-L (22k), and ConvNeXt-L). Five different models have the best score on at least one of the six datasets. The top five models' mean scores lie within 1.2 points of each other (75.6 to 76.8). Among the ten highest-ranked models, marginal 95\% confidence intervals overlap in 73\% of model-pair--dataset comparisons. This overlap is descriptive and does not establish whether paired score differences are statistically distinguishable, which would required elaborate paired significance tests.

\paragraph{Multispectral input.}
\begin{table}[t]
	\centering
	\footnotesize
	\setlength{\tabcolsep}{4pt}
\begin{tabular}{@{}lrrr@{}}
\toprule
Model & Mean$_4$ & Rank$_4$ & TreeSatAI \\
\midrule
\modelcell{GFM}{OlmoEarth v1.3 base} & 78.4 & 3.8 & -- \\
\modelcell{GFM}{TerraMind v1 large} & 77.6 & 7.0 & 64.6 \\
\modelcell{GFM}{Panopticon} & 76.1 & 9.2 & -- \\
\modelcell{GFM}{Clay v1.5 large} & 76.6 & 10.6 & 62.3 \\
\modelcell{DINOv3}{DINOv3-SAT ViT-L/16\eo} & 72.9 & 17.0 & 63.0 \\
\modelcell{GFM}{DOFA large} & 74.0 & 17.5 & 64.1 \\
\modelcell{GFM}{Prithvi-v2 300M} & 73.5 & 27.8 & 60.6 \\
\modelcell{DINOv3}{DINOv3 ViT-L/16} & 69.7 & 33.2 & 62.2 \\
\modelcell{ImageNet}{ResNet-50} & 67.9 & 51.4 & 58.6 \\
Random conv. features & 65.5 & 55.9 & 64.2 \\
Image statistics & 62.0 & 83.4 & 59.6 \\
\bottomrule
\end{tabular}

	\par\smallskip
	{\scriptsize\modelfamilylegend\quad Unshaded: no pretraining.}
	\caption{\textbf{All-band classification with a linear probe.} Mean$_4$ is the mean score ($\times100$) over BENv2, So2Sat, EuroSAT-Sp, and GEO-Bench ForestNet, and Rank$_4$ is the average rank on these four datasets among the 85 models with all-band results on all of them. TreeSatAI is shown separately because several models have no all-band result on it, and a dash marks a missing result. The rows show one checkpoint from each of six GFM families, the web-image and satellite DINOv3 ViT-L/16, and three references, sorted by Rank$_4$.}
	\label{tab:ms}
\end{table}

RGB-only results understate models built for multispectral input (\cref{tab:ms}). OlmoEarth v1.3 base with all bands reaches a mean of 78.4 on BENv2, So2Sat, EuroSAT with a spatial split (EuroSAT-Sp), and GEO-Bench ForestNet, compared with an RGB-only mean of 75.8 for DINOv3-SAT ViT-L/16, the best RGB model on these four datasets. The additional bands raise the mean score of the OlmoEarth variants by 5.1 to 12.9 points, whereas DOFA, whose wavelength-conditioned patch embedding accepts any bands, loses 1.8 points (large) and 1.6 points (base). General vision models also lose accuracy with adapted input layers, as 49 of 53 score lower with all bands, with a median change of $-2.0$ points.

\begin{figure*}[t]
	\centering
	\includegraphics[width=0.95\textwidth]{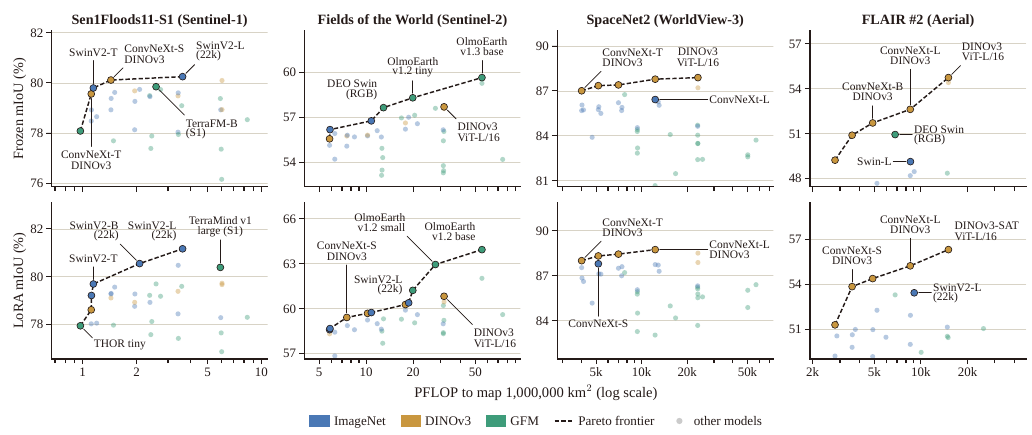}
	\caption{\textbf{mIoU against the inference FLOPs needed to map 1{,}000{,}000\,km$^2$}, with frozen (top) and LoRA (bottom) encoders. Compute is the number of tiles times the forward FLOPs of the backbone and UPerNet head per tile, at evaluation pixel sizes of 23, 10, 0.38, and 0.46\,m from left to right. These differ from the native resolution for Sen1Floods11-S1 and FLAIR \#2, because both were evaluated on inputs resized to 224 pixels. The dashed line marks the accuracy--FLOP Pareto frontier over all models with frozen and LoRA results. Scores are mean mIoU over three seeds.}
	\label{fig:mapping}
\end{figure*}

\subsection{Accuracy and cost in mapping workloads}
\label{sec:mapping}

Applications such as rapid post-disaster building damage assessment or global agricultural field mapping require processing large volumes of imagery, so inference efficiency affects how quickly maps can be delivered and how much area a fixed compute budget covers. We compare segmentation mIoU with the estimated inference compute needed to map 1{,}000{,}000\,km$^2$ ($\approx 10\%$ of the area of the US), which is the number of non-overlapping image patches needed to cover this area multiplied by the forward FLOPs of the backbone and UPerNet head. We convert FLOPs into illustrative GPU times at a common sustained rate of 25.6\,TFLOP/s, the median achieved rate of the 33 ranked models profiled at 512 pixels on one RTX PRO 6000 Blackwell GPU (supplementary \cref{app:compute}). Because achieved throughput differs across architectures, these conversions do not predict model-specific runtime.

Supplementary \cref{fig:sensor-accuracy} compares mean LoRA mIoU within each sensor group against a standardized reference cost, calculated for $224\times224$ RGB tiles at the group's nominal GSD. The accuracy scores use each dataset's evaluation configuration, which can differ in input size and band count, so these reference costs do not estimate the compute required to achieve the plotted accuracy. Across the VHR satellite datasets, ConvNeXt-S DINOv3 scores 0.7 mIoU points below DINOv3 ViT-L/16, while its reference cost is $4.2\times$ lower. On the aerial dataset, the corresponding accuracy gap is 2.5 points. We use dataset-specific input configurations and ground footprints for the mapping projections in \cref{fig:mapping}.

\Cref{fig:mapping} compares four mapping scenarios: floods (Sen1Floods11-S1), agricultural fields (FotW, Sentinel-2), buildings (SpaceNet2), and land cover (FLAIR \#2), using each dataset's evaluation resolution. With frozen encoders, flood and building mapping have cheaper models within one mIoU point of the best score. On SpaceNet2, the best frozen model requires $5.9\times$ as many inference FLOPs as ConvNeXt-T DINOv3 for a 0.9-point improvement in mIoU. At the assumed processing rate, mapping 1{,}000{,}000\,km$^2$ would take approximately 10.6 and 1.8 GPU-days, respectively. In contrast, the next cheaper frontier models on FotW and FLAIR \#2 have larger accuracy gaps. FotW is also the only scenario where OlmoEarth models lead under both frozen and LoRA adaptation, while ImageNet or DINOv3 models lead the other three. Because LoRA changes accuracy without changing projected inference cost, it also changes which models lie on the Pareto frontier.

\subsection{Sensitivity to evaluation choices}
\label{sec:choices}
To measure how evaluation choices change scores and rankings, we compare paired results across the configurations in supplementary \cref{tab:sweep} and summarize them in supplementary \cref{tab:choices}.

\paragraph{Probe type.}
Linear probes score 15.3 points higher than $k$NN on average, and the two probes rank models differently ($\tau=0.60$), although seven of the ten best models are shared. Panopticon, for example, has an average rank of 5.9 with $k$NN and 13.1 with a linear probe, whereas SwinV2-L (22k) has 32.8 with $k$NN and 15.8 with a linear probe.

\paragraph{Regularization.}
Over five models and five GEO-Bench datasets, test accuracy varies by 11.7 to 38.5 points across a separate 41-value grid of $C$ from $10^{-7}$ to $10^{2}$ (supplementary \cref{fig:csweep}). Fixing $C=1$, the scikit-learn default, costs 1.6 points on average and up to 7.8 points (ConvNeXt-L DINOv3 on So2Sat). The best value depends on the model, with a median of $10^{-2.2}$ for ConvNeXt-L DINOv3 and $10^{0.4}$ for DINOv3-SAT ViT-L/16, so any fixed value favors some models over others.

\paragraph{Probe fitting.}
We compare full-batch L-BFGS with mini-batch Adam~\cite{kingma2015adam} for ResNet-18 and DINOv3-SAT ViT-L/16 on five GEO-Bench datasets, with eight values of $C$ and, for Adam, 12 learning rates for each $C$ (supplementary \cref{subsec:solver-comparison}). Across all ten (model, dataset) pairs, recorded fitting time totals 1.4 minutes for 80 L-BFGS fits and 72.5 minutes for 960 Adam fits on one GPU. The $52\times$ reduction in total search time includes Adam's additional learning-rate search; mean recorded time per fit is approximately $4.4\times$ lower for L-BFGS. L-BFGS matches or exceeds Adam's test accuracy on 8 of the 10 pairs (means of 80.0 and 79.7).

\paragraph{Input handling.}
Normalization, and pooling under $k$NN, change scores by more than the 1.2-point gap between the five best models. Normalizing inputs with a model's pretraining statistics instead of the dataset's mean and standard deviation raises scores by 1.8 points on average over 1{,}120 (model, dataset) pairs (supplementary \cref{tab:choices}). Pooling interacts with the probe, as the class token scores 0.6 points below mean pooling with a linear probe and 4.4 points above it with $k$NN, on average over 89 pairs. Supplementary \cref{app:quirks} gives examples of both effects.

\paragraph{Segmentation protocol.}
Replacing UPerNet with UNet lowers mIoU by 0.4 points on average over 139 paired evaluations. In a separate 10-epoch comparison (supplementary \cref{sec:supp-choices}), FPN outperforms a per-pixel linear head by 3.9 points. Encoder adaptation also changes the model ordering: with the learning rates and training duration used here, LoRA and full fine-tuning produce rankings with $\tau=0.82$ and 0.69 relative to the frozen ranking across the ten models evaluated in all three regimes. On LoveDA, increasing the input size from 224 to 512 pixels improves mIoU by 3.0 points on average and up to 7.0. Across the separate duration, learning-rate, and schedule variants (supplementary \cref{sec:supp-choices}), the mean change relative to the 10-epoch reference is $-0.3$ points, although individual (model, dataset) pairs change by up to 11.1 points.

\subsection{Uncertainty}
\label{sec:reliability}
In our experiments the median 95\% bootstrap half-width ranges from 0.7 points on EuroSAT-spatial, with 5,400 test images, to 3.0 points on GEO-Bench ForestNet, with 993. These intervals use 200 resamples of the test predictions. As such, they quantify test-sampling uncertainty conditional on that fit and split, excluding variation from training-data sampling and validation-based selection of the regularization parameter. Because models share the same test examples, overlapping marginal intervals can coexist with a paired difference whose interval excludes zero, and does not establish statistical equivalence or significance. For segmentation, the standard deviation of mIoU across three training seeds has a median of 0.4 and a 90th percentile of 1.2 points. This measures between-run variation on the fixed benchmark split, rather than test-sampling uncertainty.

\section{Discussion}
\label{sec:discussion}

\paragraph{Recommendations.}
Results from a shared harness can be compared across tasks, input settings, adaptation methods, and inference costs, which tells a user more than an aggregate rank. We recommend reporting confidence intervals and inference compute next to accuracy, and we encourage others to add results with the same pipeline and to publish the configuration with each score. For segmentation we ask for three training seeds per configuration, and for classification the harness already runs both probes, the regularization search, and the bootstrap intervals.

\paragraph{Limitations.}
Our benchmark datasets do not capture the diversity of conditions that impact spatial and sensor generalization. Model rankings may differ where geography and label distributions are different, or where input distributions shift due to sensor conditions. Our mapping-time projections also assume one processing rate for all models, so they do not capture model-specific hardware efficiency or end-to-end runtime. Apart from the small segmentation comparison in \cref{sec:choices}, we do not evaluate full fine-tuning, because fine-tuning every model on every dataset would be extremely computationally expensive. Finally, although we evaluate more than 100 models, they are not a complete catalog of GFMs. Our benchmark datasets also cannot yet test models such as TESSERA~\cite{feng2025tessera}, which embed a time series of Sentinel-2 and Sentinel-1 radiometrically terrain-corrected (RTC) observations per pixel, because our core datasets evaluation focuses on single time steps.The harness currently also does not provide functionality to evaluate downstream performance on various tasks through retrieved published embeddings from models like AlphaEarth \cite{brown2025alphaearth}.

\paragraph{Future work.}
Benchmarks need harder datasets. In early experiments, most models scored above 95\% on PV4GER and Brick Kiln (\cref{sec:suite}), so these datasets no longer separate models. Which subset of datasets to compare models on is also an open question, because the choice changes aggregate rankings~\cite{dehghani2021benchmarklottery,zhang2024inherent}. New tasks should also reduce the geographic bias of current datasets and/or score predictions by the cost of the decisions they inform. In deforestation monitoring, for example, a false positive could send investigators out for nothing.

Our results also expose gaps in model design. On the four aerial and very-high-resolution optical datasets in \cref{tab:seg-lora}, the best GFM scores 1.5 to 4.0 mIoU below the best ImageNet or DINOv3 model, which motivates GFMs designed for high-resolution RGB imagery. \citet{sharma2026geofms} report a similar result on ten VHR datasets, where encoders pretrained on natural images or VHR imagery transfer better than those pretrained only on Sentinel-scale imagery. Most GFMs also release only a pretrained encoder, so every dense-prediction task trains its decoder from scratch. Pretraining decoders together with the encoder, or adapting model-agnostic upsamplers such as AnyUp~\cite{wimmer2025anyup} to EO data, could reduce this task-specific training. Many applications also rely on time series, and how GFMs can use them efficiently is not well understood. The harness also records calibration error and intrinsic-dimension estimates, which we do not analyze here. Relating them to transfer performance~\cite{lehmann2026calibration,rao2026measuring} could help explain why some representations transfer better than others.

\section{Conclusion}
\label{sec:conclusion}

We present \texttt{geo-eval-harness}, a shared harness that evaluates every model with the same open-source code and stores the configuration of each run. Using it, we find that evaluation choices can change scores by more than the leading models differ, that the benefit of EO pretraining depends on the imagery and task, and that similarly accurate models can require substantially different inference FLOPs for large-area mapping. At the assumed processing rate, these differences can amount to days of GPU time. We release the code, configurations, and results for others to reproduce and extend.

\section*{Acknowledgements}
We thank Gabriel Tseng, Anthony Fuller, Hamed Alemohammad, Evan Shelhamer, and Konstantin Klemmer for helpful discussions.

We also thank the model authors who reviewed our implementations and results in the \texttt{geo-eval-harness} repository: Ritwik Gupta (Scale-MAE), Anthony Fuller (CROMA), Johannes Jakubik and Benedikt Blumenstiel (TerraMind and Prithvi), Daniela Szwarcman (Prithvi), Leonard Waldmann (Panopticon), Filip Wolf and Bla\v{z} Rolih (DEO), Guillaume Astruc (UniverSat), Gabriel Tseng (OlmoEarth), Piper Wolters (OlmoEarth and SatlasNet), Muhammad Sohail Danish (TerraFM), Bruno S\'anchez-Andrade Nu\~no (Clay), Alexandre Lacoste (SeCo), and Yi Wang (Copernicus-FM). Their comments identified problems in how we preprocessed inputs, normalized bands, and extracted features for several models. Some of them also contributed fixes, updated checkpoints, and datasets to the harness.

This research used computing resources provided by the Microsoft AI for Good lab as well as the TGI RAILs advanced compute and data resource which is supported by the National Science Foundation (award OAC-2232860) and Taylor Geospatial.

We used LLM-based tools (GitHub Copilot and Claude) to assist with this work. Specifically, we used LLMs in developing \texttt{geo-eval-harness}, the scripts that generate our tables and figures, and reviewing code changes -- with manual review by the authors. We further used LLMs to proofread and revise the paper. The ideas and aims of the paper are the authors' own, and the authors take full responsibility for its content.

{
	\small
	\bibliographystyle{ieeenat_fullname}
	\bibliography{main}
}

\onecolumn
\iftoggle{cvprfinal}{}{\nolinenumbers}
\setcounter{page}{1}
\renewcommand{\thepage}{S\arabic{page}}
\begin{center}
  \Large
  \textbf{\thetitle}\\[0.5em]
  Supplementary Material
\end{center}
\medskip

\setcounter{section}{0}
\setcounter{table}{0}
\setcounter{figure}{0}
\setcounter{lstlisting}{0}
\renewcommand{\thesection}{S\arabic{section}}
\renewcommand{\thetable}{S\arabic{table}}
\renewcommand{\thefigure}{S\arabic{figure}}
\renewcommand{\thelstlisting}{S\arabic{lstlisting}}
\renewcommand{\theHsection}{S\arabic{section}}
\renewcommand{\theHtable}{S\arabic{table}}
\renewcommand{\theHfigure}{S\arabic{figure}}

\iftoggle{cvprfinal}{%
  \newenvironment{nolinenumbers}{}{}%
}{%
  \runninglinenumbers[1]
  \leftlinenumbers
  \setlength{\linenumbersep}{4pt}
}

\raggedbottom
\newenvironment{supplementtable}{%
  \par\medskip\begin{nolinenumbers}\noindent
  \begin{minipage}{\linewidth}\captionsetup{type=table}\centering
}{\end{minipage}\par\end{nolinenumbers}\medskip}
\newenvironment{supplementfigure}{%
  \par\medskip\begin{nolinenumbers}\noindent
  \begin{minipage}{\linewidth}\captionsetup{type=figure}\centering
}{\end{minipage}\par\end{nolinenumbers}\medskip}
\newcommand{\noterule}{\arrayrulecolor{black!25}\specialrule{0.3pt}{2pt}{2pt}\arrayrulecolor{black}}

This supplement lists the datasets and models, describes how we normalize inputs, measure compute, and select and rank results, and reports the comparisons behind \cref{sec:choices}. Per-model scores on every dataset are in the released results.

\section{Datasets}
\label{sec:supp-datasets}

\Cref{tab:supp-datasets} lists all datasets in \texttt{geo-eval-harness} (6 classification and 17 segmentation) with their sources, harness names, sensors, band counts, and splits. The 14 core segmentation datasets are all of these except the single-band BRIGHT task, LEVIR-CD, and Open Cities AI. GEO-Bench datasets use the default partitions that come packaged with the datasets. EuroSAT-Sp uses the fixed 60/20/20 spatial split provided by TorchGeo's EuroSATSpatial dataset~\citep{stewart2022torchgeo}. For LoveDA, we hold out 10\% of the official training tiles within each Urban/Rural group for validation and use the official validation set as our test set.

\begin{supplementtable}
	\centering
	\scriptsize
	\setlength{\tabcolsep}{4pt}
\begin{tabular}{@{}llllrrlrr@{}}
\toprule
Dataset & Harness name & Task & Sensor & Bands & Cls. & Train / Val / Test & Metric & Models \\
\midrule
BigEarthNet v2 (GEO-Bench-2)~\cite{clasen2025reben,simumba2026geobench} & \texttt{benv2} & ML & S2+S1 & 14 & 19 & 20,000 / 4,000 / 4,000 & mAP & 105 \\
So2Sat LCZ42 (GEO-Bench-2)~\cite{zhu2020so2sat,simumba2026geobench} & \texttt{so2sat} & SL & S2+S1 & 12 & 17 & 19,992 / 986 / 986 & Acc. & 105 \\
EuroSAT, spatial split~\cite{helber2019eurosat} & \texttt{eurosat-spatial} & SL & S2 & 13 & 10 & 16,200 / 5,400 / 5,400 & Acc. & 105 \\
ForestNet (GEO-Bench)~\cite{irvin2020forestnet,lacoste2023geobench} & \texttt{m-forestnet} & SL & Landsat & 6 & 12 & 6,464 / 989 / 993 & Acc. & 101 \\
TreeSatAI (GEO-Bench-2)~\cite{ahlswede2023treesatai,simumba2026geobench} & \texttt{treesatai} & ML & Aerial+S2+S1 & 19 & 15 & 4,000 / 1,000 / 2,000 & mAP & 100 \\
NWPU-RESISC45~\cite{cheng2017resisc45} & \texttt{resisc45} & SL & Aerial & 3 & 45 & 18,900 / 6,300 / 6,300 & Acc. & 104 \\
\midrule
BRIGHT post-event SAR~\cite{chen2025bright} & \texttt{bright-post-sar} & Seg & SAR & 1 & 4 & 2,918 / 450 / 878 & mIoU & 32 \\
BRIGHT post-event SAR, 3 channels~\cite{chen2025bright} & \texttt{bright-post-sar-rgb} & Seg & SAR (3 ch.) & 3 & 4 & 2,918 / 450 / 878 & mIoU & 49 \\
HLS Burn Scars (GEO-Bench-2)~\cite{jakubik2023prithvi,simumba2026geobench} & \texttt{burn\_scars} & Seg & S2 & 6 & 3 & 524 / 160 / 120 & mIoU & 53 \\
CaFFe (GEO-Bench-2)~\cite{gourmelon2022caffe,simumba2026geobench} & \texttt{caffe} & Seg & SAR (gray) & 3 & 4 & 4,000 / 1,000 / 2,000 & mIoU & 49 \\
CloudSEN12 (GEO-Bench-2)~\cite{aybar2022cloudsen12,simumba2026geobench} & \texttt{cloudsen12} & Seg & S2 & 12 & 4 & 4,000 / 535 / 975 & mIoU & 58 \\
FLAIR \#2 (GEO-Bench-2)~\cite{garioud2023flair,simumba2026geobench} & \texttt{flair2} & Seg & Aerial+DEM & 5 & 13 & 4,049 / 1,022 / 3,022 & mIoU & 49 \\
Fields of The World (GEO-Bench-2)~\cite{kerner2025fotw,simumba2026geobench} & \texttt{fotw} & Seg & S2 & 4 & 3 & 4,000 / 1,000 / 2,000 & mIoU & 47 \\
HOTOSM buildings~\cite{hotosm2025buildings} & \texttt{hotosm\_buildings} & Seg & Aerial & 3 & 2 & 5,000 / 7,237 / 7,236 & mIoU & 49 \\
Kuro Siwo (GEO-Bench-2)~\cite{bountos2024kurosiwo,simumba2026geobench} & \texttt{kuro\_siwo} & Seg & S1+DEM & 3 & 4 & 4,000 / 1,000 / 2,000 & mIoU & 38 \\
LEVIR-CD, post-change image~\cite{chen2020levircd} & \texttt{levir-cd-post} & Seg & Aerial & 3 & 2 & 7,120 / 1,024 / 2,048 & mIoU & 50 \\
LoveDA~\cite{wang2021loveda} & \texttt{loveda} & Seg & Aerial & 3 & 7 & 2,269 / 253 / 1,669 & mIoU & 47 \\
OpenEarthMap-SAR~\cite{xia2025oemsar} & \texttt{openearthmap-sar} & Seg & SAR & 1 & 8 & 4,333 / 210 / 490 & mIoU & 43 \\
Open Cities AI~\cite{gfdrr2020opencities} & \texttt{open\_cities} & Seg & Aerial & 3 & 2 & 10,000 / 11,236 / 22,408 & mIoU & 49 \\
Sen1Floods11, Sentinel-1~\cite{bonafilia2020sen1floods11} & \texttt{sen1floods11-s1} & Seg & S1 & 2 & 2 & 252 / 89 / 90 & mIoU & 47 \\
Sen1Floods11, Sentinel-2~\cite{bonafilia2020sen1floods11} & \texttt{sen1floods11-s2} & Seg & S2 & 13 & 2 & 252 / 89 / 90 & mIoU & 56 \\
SpaceNet 2 (GEO-Bench-2)~\cite{vanetten2018spacenet,simumba2026geobench} & \texttt{spacenet2} & Seg & WorldView+Pan & 9 & 2 & 5,186 / 1,461 / 2,961 & mIoU & 47 \\
SpaceNet 7 (GEO-Bench-2)~\cite{vanetten2021spacenet7,simumba2026geobench} & \texttt{spacenet7} & Seg & PlanetScope & 3 & 2 & 3,500 / 652 / 1,152 & mIoU & 47 \\
\bottomrule
\end{tabular}

	\caption{\textbf{\texttt{geo-eval-harness} Datasets.} Task is single-label (SL) or multi-label (ML) classification, or semantic segmentation (Seg). Bands is the number of channels with all bands selected, and Cls.\ is the number of classes. Models is the number of models with an RGB linear-probe result (classification) or a frozen UPerNet result (segmentation) on the dataset.}
	\label{tab:supp-datasets}
\end{supplementtable}

\section{Models}
\label{sec:supp-models}

\Cref{tab:supp-models} lists the 100 models in \texttt{geo-eval-harness} that we rank in RGB classification, ordered by their average rank with a linear probe. Parameters and GFLOPs are for the backbone, computed with a $224\times224$ RGB input (\cref{app:compute}). Throughput is classification inference at a batch size of 1 (\cref{app:compute}). A dash marks a model with no all-band rank or no frozen UPerNet segmentation results.

\begin{supplementtable}
	\centering
	\footnotesize
	\setlength{\tabcolsep}{6pt}
\begin{tabular}{@{}llrrrrrrr@{}}
\toprule
Model & Pretrain & Params (M) & GFLOPs & img/s & Lin. & $k$NN & MS$_4$ & Seg \\
\midrule
DOFA large & EO & 337.2 & 123.5 & 132 & 3.8 & 9.5 & 17.5 & 17 \\
DINOv3 ViT-L/16 & Natural & 303.1 & 125.7 & 116 & 4.4 & 11.8 & 33.2 & 17 \\
DEO Swin (RGB) & EO & 86.9 & 51.0 & 83 & 4.9 & 3.7 & -- & 17 \\
DINOv3-SAT ViT-L/16 & EO & 303.1 & 125.7 & 123 & 5.2 & 6.8 & 17.0 & 17 \\
ViT-L/16 & Natural & 303.3 & 123.1 & 145 & 7.9 & 12.6 & 20.1 & 17 \\
\addlinespace
DOFA base & EO & 111.3 & 35.4 & 268 & 10.1 & 16.8 & 23.5 & 17 \\
Panopticon & EO & 98.1 & 64.0 & 213 & 13.1 & 5.9 & 9.2 & 16 \\
Swin-L & Natural & 195.0 & 68.9 & 85 & 15.7 & 38.1 & 33.4 & 17 \\
SwinV2-L (22k) & Natural & 195.2 & 74.1 & 76 & 15.8 & 32.8 & 25.4 & 17 \\
ConvNeXt-L & Natural & 196.2 & 68.7 & 185 & 16.7 & 28.5 & 22.0 & 17 \\
\addlinespace
ConvNeXt-B & Natural & 87.6 & 30.7 & 209 & 18.5 & 24.8 & 27.4 & 17 \\
TerraMind v1 base & EO & 85.5 & 34.9 & 336 & 22.8 & 28.1 & 16.2 & 16 \\
EfficientNet-B4 & Natural & 17.5 & 3.0 & 134 & 23.0 & 37.6 & 45.5 & 16 \\
TerraMind v1 large & EO & 302.8 & 122.4 & 140 & 23.8 & 35.9 & 7.0 & 16 \\
Clay v1.5 large & EO & 311.4 & 535.2 & 58 & 25.1 & 53.9 & 10.6 & -- \\
\addlinespace
SwinV2-B (22k) & Natural & 86.9 & 34.1 & 75 & 25.2 & 43.5 & 40.9 & 17 \\
DeiT-B/16 & Natural & 85.8 & 35.1 & 326 & 27.1 & 19.7 & 33.1 & -- \\
ConvNeXtV2-B & Natural & 87.7 & 30.7 & 136 & 27.8 & 37.8 & 35.0 & -- \\
Swin-B & Natural & 86.7 & 30.9 & 81 & 28.1 & 47.6 & 45.8 & 17 \\
ConvNeXt-L DINOv3 & Natural & 196.2 & 68.7 & 187 & 28.7 & 34.1 & 43.8 & 17 \\
\addlinespace
MoCo ResNet-50 (RGB) & EO & 23.5 & 8.2 & 326 & 28.8 & 19.8 & -- & -- \\
ConvNeXt-B DINOv3 & Natural & 87.6 & 30.7 & 219 & 28.9 & 44.6 & 53.9 & 17 \\
ConvNeXt-T DINOv3 & Natural & 27.8 & 8.9 & 403 & 29.2 & 53.1 & 54.0 & 17 \\
ScaleMAE large & EO & 304.3 & 123.1 & 137 & 30.2 & 33.2 & -- & 9 \\
ConvNeXt-T & Natural & 27.8 & 8.9 & 402 & 30.3 & 51.0 & 39.1 & 17 \\
\addlinespace
Swin-S & Natural & 48.8 & 17.5 & 86 & 33.0 & 54.1 & 54.2 & 17 \\
DenseNet-161 & Natural & 26.5 & 15.5 & 98 & 34.6 & 25.0 & 49.2 & 17 \\
SeCo ResNet-50 & EO & 23.5 & 8.2 & 334 & 34.9 & 31.9 & -- & -- \\
DenseNet-169 & Natural & 12.5 & 6.7 & 91 & 36.8 & 18.2 & 47.2 & -- \\
ConvNeXt-S DINOv3 & Natural & 49.5 & 17.4 & 218 & 36.9 & 53.8 & 47.5 & 17 \\
\addlinespace
ViT-B/16 & Natural & 85.8 & 35.1 & 327 & 37.4 & 29.8 & 44.0 & 17 \\
DenseNet-201 & Natural & 18.1 & 8.6 & 79 & 38.0 & 25.8 & 53.9 & -- \\
OlmoEarth v1 base & EO & 89.0 & 604.4 & 51 & 38.7 & 38.0 & 7.4 & -- \\
ConvNeXtV2-T & Natural & 27.9 & 8.9 & 262 & 39.8 & 43.3 & 46.9 & -- \\
OlmoEarth v1.2 base & EO & 114.0 & 160.8 & 97 & 39.8 & 40.8 & 7.4 & 16 \\
\addlinespace
DeiT-S/16 & Natural & 21.7 & 9.2 & 405 & 40.6 & 24.8 & 51.1 & -- \\
Swin-T & Natural & 27.5 & 9.0 & 168 & 40.8 & 48.8 & 46.8 & 17 \\
EfficientNet-B3 & Natural & 10.7 & 1.9 & 160 & 40.8 & 40.1 & 59.2 & -- \\
WRN-50-2 & Natural & 66.8 & 22.8 & 298 & 41.1 & 34.0 & 51.9 & -- \\
SwinV2-B & Natural & 86.9 & 33.8 & 57 & 41.8 & 51.2 & 48.9 & -- \\
\addlinespace
Clay v1 base & EO & 92.1 & 156.5 & 133 & 43.0 & 87.3 & 13.4 & 9 \\
SatlasNet SwinV2-B (MI) & EO & 86.9 & 34.1 & 59 & 44.1 & 65.4 & -- & -- \\
SwinV2-T & Natural & 27.6 & 9.7 & 119 & 45.2 & 52.5 & 41.2 & 17 \\
MobileNetV3-L & Natural & 4.2 & 0.4 & 331 & 46.0 & 43.6 & 46.2 & 16 \\
ResNet-50 & Natural & 23.5 & 8.2 & 345 & 46.5 & 33.5 & 51.4 & 16 \\
\addlinespace
DenseNet-121 & Natural & 7.0 & 5.7 & 129 & 46.8 & 38.8 & 41.0 & -- \\
SatlasNet SwinV2-B NAIP (SI) & EO & 86.9 & 34.1 & 61 & 46.8 & 50.7 & -- & -- \\
SatlasNet SwinV2-B NAIP (MI) & EO & 86.9 & 34.1 & 58 & 47.2 & 50.7 & -- & -- \\
SwinV2-S & Natural & 49.0 & 19.2 & 58 & 48.0 & 57.4 & 53.0 & 17 \\
Prithvi-v2 600M & EO & 630.4 & 334.6 & 59 & 48.0 & 88.7 & 31.0 & 13 \\
\bottomrule
\end{tabular}

	\caption{\textbf{Models}, part 1 of 2. Pretrain is EO data, natural images, or none. Lin.\ and $k$NN are average ranks on the six core datasets with RGB input, among the same 100 models. MS$_4$ is the average all-band rank on BENv2, So2Sat, EuroSAT-Sp, and GEO-Bench ForestNet among 85 models. Seg is the number of segmentation datasets with frozen UPerNet results.}
	\label{tab:supp-models}
\end{supplementtable}

\begin{supplementtable}
	\ContinuedFloat
	\centering
	\footnotesize
	\setlength{\tabcolsep}{6pt}
\begin{tabular}{@{}llrrrrrrr@{}}
\toprule
Model & Pretrain & Params (M) & GFLOPs & img/s & Lin. & $k$NN & MS$_4$ & Seg \\
\midrule
ViT-S/16 & Natural & 21.7 & 9.2 & 396 & 48.0 & 32.8 & 36.9 & 17 \\
OlmoEarth v1 large & EO & 307.8 & 1965.6 & 21 & 48.2 & 43.4 & 7.1 & -- \\
SeCo ResNet-18 & EO & 11.2 & 3.6 & 768 & 48.7 & 34.2 & -- & -- \\
ConvNeXt-S & Natural & 49.5 & 17.4 & 211 & 49.0 & 53.6 & 63.5 & 17 \\
ResNet-101 & Natural & 42.5 & 15.6 & 171 & 49.3 & 35.8 & 52.8 & 17 \\
\addlinespace
RegNetY-800MF & Natural & 5.5 & 1.6 & 235 & 50.3 & 42.6 & 51.8 & -- \\
MoCo ResNet-18 (RGB) & EO & 11.2 & 3.6 & 801 & 51.5 & 23.9 & -- & -- \\
OlmoEarth v1.3 base & EO & 128.3 & 187.0 & 77 & 51.7 & 41.8 & 3.8 & 16 \\
OlmoEarth v1.1 base & EO & 114.0 & 160.8 & 107 & 52.1 & 44.3 & 17.8 & -- \\
ResNet-152 & Natural & 58.1 & 23.0 & 115 & 54.8 & 40.8 & 58.4 & -- \\
\addlinespace
SatlasNet SwinV2-B (SI) & EO & 86.9 & 34.1 & 62 & 54.8 & 51.1 & -- & -- \\
OlmoEarth v1.2 small & EO & 35.6 & 47.2 & 100 & 54.9 & 47.8 & 5.0 & 16 \\
MobileNetV3-S & Natural & 1.5 & 0.1 & 379 & 56.6 & 54.1 & 56.4 & -- \\
SatlasNet SwinV2-T (MI) & EO & 27.6 & 9.9 & 118 & 56.6 & 69.3 & -- & -- \\
RegNetX-200MF & Natural & 2.3 & 0.4 & 379 & 60.1 & 52.3 & 68.5 & -- \\
\addlinespace
Prithvi-v2 300M & EO & 303.1 & 123.1 & 139 & 60.1 & 88.8 & 27.8 & 13 \\
RegNetX-800MF & Natural & 6.6 & 1.6 & 318 & 61.2 & 48.3 & 67.2 & -- \\
MaxViT-T & Natural & 30.4 & 11.1 & 69 & 62.3 & 64.7 & 70.4 & -- \\
RegNetY-200MF & Natural & 2.8 & 0.4 & 263 & 63.2 & 38.8 & 57.8 & -- \\
GASSL ResNet-50 & EO & 23.5 & 8.2 & 332 & 63.4 & 62.1 & -- & -- \\
\addlinespace
SMARTIES large & EO & 307.0 & 123.1 & 142 & 63.7 & 76.2 & -- & 16 \\
ResNet-18 & Natural & 11.2 & 3.6 & 797 & 64.9 & 47.1 & 74.2 & 17 \\
Prithvi-v2 300M (TL) & EO & 303.1 & 123.1 & 139 & 65.0 & 85.8 & 28.8 & 13 \\
SatlasNet SwinV2-T (SI) & EO & 27.6 & 9.9 & 121 & 67.2 & 63.8 & -- & -- \\
SMARTIES base & EO & 88.6 & 35.1 & 314 & 67.4 & 79.9 & -- & 16 \\
\addlinespace
DeiT-Ti/16 & Natural & 5.5 & 2.5 & 408 & 70.1 & 40.3 & 71.4 & -- \\
ResNet-34 & Natural & 21.3 & 7.3 & 455 & 72.6 & 51.3 & 73.1 & -- \\
ViT-Ti/16 & Natural & 5.5 & 2.5 & 403 & 72.7 & 42.8 & 64.2 & -- \\
VGG-16 & Natural & 134.3 & 30.9 & 489 & 72.8 & 74.2 & 71.4 & -- \\
Prithvi-v2 100M (TL) & EO & 85.6 & 35.1 & 305 & 72.9 & 86.7 & 27.0 & 13 \\
\addlinespace
UniverSat base & EO & 201.1 & 115616.1 & 0 & 75.0 & 76.3 & 46.5 & 1 \\
EfficientNet-B0 & Natural & 4.0 & 0.8 & 259 & 77.3 & 63.2 & 67.5 & -- \\
VGG-19 & Natural & 139.6 & 39.3 & 450 & 77.3 & 72.5 & 75.0 & -- \\
SatlasNet R-152 (SI) & EO & 58.1 & 23.0 & 112 & 77.8 & 70.9 & -- & 16 \\
OlmoEarth v1 tiny & EO & 6.2 & 76.2 & 114 & 78.4 & 65.0 & 17.2 & -- \\
\addlinespace
EfficientNet-B2 & Natural & 7.7 & 1.3 & 174 & 78.8 & 66.2 & 74.6 & 15 \\
OlmoEarth v1.1 tiny & EO & 12.5 & 15.3 & 154 & 81.9 & 73.2 & 18.5 & -- \\
OlmoEarth v1.2 tiny & EO & 12.5 & 15.3 & 101 & 82.2 & 64.8 & 20.1 & 16 \\
SatlasNet R-152 (MI) & EO & 58.1 & 23.0 & 114 & 86.1 & 78.5 & -- & -- \\
Prithvi-v1 100M & EO & 85.6 & 35.1 & 302 & 87.3 & 90.0 & 43.0 & 13 \\
\addlinespace
EfficientNetV2-S & Natural & 20.2 & 5.7 & 115 & 88.8 & 78.2 & 74.0 & -- \\
OlmoEarth v1 nano & EO & 1.4 & 15.2 & 213 & 89.2 & 75.9 & 38.2 & -- \\
SatlasNet R-50 (MI) & EO & 23.5 & 8.2 & 318 & 90.2 & 84.3 & -- & -- \\
SatlasNet R-50 (SI) & EO & 23.5 & 8.2 & 320 & 91.2 & 81.5 & -- & -- \\
EfficientNetV2-M & Natural & 52.9 & 10.7 & 82 & 91.8 & 76.8 & 69.5 & -- \\
\addlinespace
OlmoEarth v1.2 nano & EO & 1.7 & 2.7 & 179 & 92.8 & 81.2 & 25.1 & 16 \\
EfficientNetV2-L & Natural & 117.2 & 24.5 & 60 & 94.2 & 85.5 & 70.0 & -- \\
OlmoEarth v1.1 nano & EO & 1.7 & 2.7 & 236 & 95.8 & 90.8 & 38.8 & -- \\
Random conv. features & None & 0.0 & 1.4 & 2,839 & 96.5 & 99.0 & 55.9 & -- \\
Image statistics & None & 0.0 & 0.0 & 18,225 & 99.7 & 98.0 & 83.4 & -- \\
\bottomrule
\end{tabular}

	\caption{\textbf{Models}, part 2 of 2.}
\end{supplementtable}

\section{Additional details}

\Cref{tab:settings} lists the evaluation settings, and \cref{lst:wrappers} shows the model and dataset interfaces described in \cref{sec:wrappers}.

\begin{supplementtable}
	\centering
	\footnotesize
	\setlength{\tabcolsep}{3pt}
	\begin{tabularx}{\linewidth}{@{}l>{\raggedright\arraybackslash}X>{\raggedright\arraybackslash}X@{}}
		\toprule
		Setting                  & Options                                       & Main results                                                                 \\
		\midrule
		\multicolumn{3}{@{}l}{\textit{Input}}                                                                                                                   \\
		Bands                    & RGB, all, or named bands                      & RGB; all in \cref{tab:ms}; per model in segmentation (\cref{sec:supp-rules}) \\
		Normalization & \texttt{dataset}, \texttt{model}, \texttt{minmax}, \texttt{minmax\_zscore}, \texttt{none} & \texttt{dataset} or \texttt{model}, per model (\cref{sec:supp-rules}) \\             
		Image size    & any                             & 224 by default; model and task exceptions \\
        Interpolation & bilinear, bicubic, area, nearest & bilinear by default; model exceptions \\
        Token pooling & mean, class token, both, model default & mean by default; model-specific embedding sources \\              
		Training subset          & GEO-Bench partitions                          & full                                                                         \\
		\midrule
		\multicolumn{3}{@{}l}{\textit{Classification}}                                                                                                          \\
		Probe                    & $k$NN, linear                                 & both                                                                         \\
		Neighbors                & any $k$                                       & 5                                                                            \\
		$C$ grid                 & range and count                               & 40 values, $10^{-6}$ to $10^{4}$                                             \\
		Refit on train and val   & yes, no                                       & yes                                                                          \\
		Feature standardization  & yes, no                                       & no                                                                           \\
		Bootstrap resamples      & any                                           & 200                                                                          \\
		\midrule
		\multicolumn{3}{@{}l}{\textit{Segmentation}}                                                                                                            \\
		Head                     & linear, conv.\ block, FPN, DPT, UNet, UPerNet & UPerNet                                                                      \\
		ViT neck                 & none, simple feature pyramid                  & pyramid for plain ViTs                                                       \\
		Adaptation               & frozen, LoRA, full fine-tuning                & frozen and LoRA; full in \cref{sec:choices}                                  \\
		LoRA                     & rank, scale, target layers                    & rank 8, $\alpha=16$, attention and MLP                                       \\
		Head learning rate       & any                                           & $10^{-3}$                                                                    \\
		Encoder learning rate    & any                                           & $10^{-4}$ or $5\times10^{-4}$ (LoRA), $10^{-5}$ (full)                       \\
		Schedule                 & epochs; cosine or constant                    & 30 epochs, cosine                                                            \\
		Weight decay, batch size & any                                           & 0, 16                                                                        \\
		Seeds                    & any                                           & 3                                                                            \\
		\bottomrule
	\end{tabularx}
	\caption{\textbf{Evaluation settings.} The table shows settings that the harness supports and their values used in the main results of the paper's experiments. \Cref{tab:app-models} lists the model families whose input handling differs from these values.}
	\label{tab:settings}
\end{supplementtable}

\par\medskip
\begin{nolinenumbers}
\noindent\begin{minipage}{\linewidth}
\captionsetup{type=lstlisting}
\begin{lstlisting}[style=harness,basicstyle=\ttfamily\footnotesize,caption={Model and dataset wrappers. The harness supplies the bands, normalizes inputs, and runs extraction, probing, and logging.},label=lst:wrappers]
class MyModel(BenchModel):
    def __init__(self, bands, *, normalization):
        super().__init__(bands, normalization)
        self.wl = [b.wavelength_um for b in bands]
        self.net = load_pretrained_backbone()

    def _forward_patch_features(self, images):
        return self.net(images, self.wl)  # (B, K)

class MyDataset(BenchDataset):
    name, task = "my_dataset", "classification"
    num_classes, multilabel = 10, False
    class_names = tuple(str(i) for i in range(10))
    rgb_bands = ["red", "green", "blue"]
    split_sizes = {"train": 2000, "val": 1000, "test": 1000}
    bands = [
        BandSpec("s2", "red", "B04", mean=957.3, std=592.8, min=0, max=28000, wavelength_um=0.665),
        ...
    ]

    def get_dataset(self, split, *, partition="default", bands=None, transform=None):
        ...  # {"image": raw float32, "label": y}
\end{lstlisting}
\end{minipage}
\par\end{nolinenumbers}\medskip

\subsection{Evaluation configurations}
Our experiments sweep the input, the classification probe, and the segmentation setup along the dimensions in \cref{tab:sweep}. The configurations default to using RGB bands, \texttt{dataset} normalization, a $224{\times}224$ bilinear resize, and seed 0. Each experiment varies a subset of these dimensions.

\begin{supplementtable}
	\centering
	\footnotesize
	\setlength{\tabcolsep}{8pt}
	\begin{tabular}{@{}ll@{}}
		\toprule
		Dimension      & Values                                                      \\
		\midrule
		\multicolumn{2}{@{}l}{\emph{Input (both tasks)}}                             \\
		Bands          & RGB; all; SAR; named subsets                                \\
		Normalization  & \texttt{dataset}; \texttt{model}; \texttt{none}             \\
		Image size     & $\{224,256,512\}$; native                                   \\
		Interpolation  & bilinear; area                                              \\
		\midrule
		\multicolumn{2}{@{}l}{\emph{Classification}}                                 \\
		Probe          & $k$NN ($k=5$); logistic regression                          \\
		Regularization & $40$ log-spaced $C \in [10^{-6},10^4]$, chosen on validation \\
		Pooling        & model default; class token; mean                            \\
		\midrule
		\multicolumn{2}{@{}l}{\emph{Segmentation}}                                   \\
		Decoder        & UPerNet; UNet; FPN; linear; conv.\ block; DPT               \\
		Adaptation     & frozen; LoRA ($r=8$); full                                  \\
		Encoder LR     & LoRA: $\{10^{-4},5\times10^{-4}\}$; full: $10^{-5}$         \\
		Neck           & none; ViT neck                                              \\
		Epochs         & $\{10,30,50\}$                                         \\
		Seeds          & $\{0,1,2\}$                                                 \\
		\bottomrule
	\end{tabular}
	\caption{\textbf{Evaluation configurations.} Settings and the values that our experiments cover for classification and segmentation tasks.}
	\label{tab:sweep}
\end{supplementtable}

\subsection{Normalization}
\label{sec:supp-normalization}

Proper data normalization ensures that the pixel value ranges of inputs a model receives is aligned with its pretrained normalization regime  (\cref{sec:wrappers}).

\begin{supplementfigure}
	\centering
	\begin{minipage}{0.55\textwidth}
\begingroup
\definecolor{nzcoral}{HTML}{FF4F2C}
\definecolor{nzblue}{HTML}{4A78B5}
\definecolor{nzink}{HTML}{1A1A1A}
\definecolor{nzsoft}{HTML}{D8D3C6}
\newcommand{\nzsize}[2]{\fontfamily{ptm}\fontsize{#1}{#2}\selectfont}
\newcommand{\nzbump}[5]{%
  \pgfmathsetmacro{\nza}{X(#1)}\pgfmathsetmacro{\nzb}{X(#2)}%
  \pgfmathsetmacro{\nzc}{(\nza+\nzb)/2}\pgfmathsetmacro{\nzw}{\nzb-\nza}%
  \path[#4] (\nza,#3) .. controls +({0.30*\nzw},0) and +({-0.20*\nzw},0) .. (\nzc,{#3+5.2})
    .. controls +({0.20*\nzw},0) and +({-0.30*\nzw},0) .. (\nzb,#3) -- cycle;
  \path[#5] (\nza,#3) .. controls +({0.30*\nzw},0) and +({-0.20*\nzw},0) .. (\nzc,{#3+5.2})
    .. controls +({0.20*\nzw},0) and +({-0.30*\nzw},0) .. (\nzb,#3);}
\newcommand{\nzcheck}[1]{\draw[nzink, line width=0.65pt, line cap=round, line join=round]
  #1 ++(-1.25,0.05) -- ++(0.75,-0.8) -- ++(1.35,1.75);}
\newcommand{\nzcross}[1]{\draw[nzink!70, line width=0.6pt, line cap=round]
  #1 ++(-0.85,-0.85) -- ++(1.7,1.7) #1 ++(-0.85,0.85) -- ++(1.7,-1.7);}
\newcommand{\nzpow}[1]{10\textsuperscript{#1}}
\newcommand{\nzdiv}{\textdiv\kern0.08em}
\newcommand{\nzx}{\texttimes}
\begin{tikzpicture}[x=0.01\linewidth, y=0.01\linewidth, text=nzink,
  font=\nzsize{7}{8.4},
  declare function={X(\d)=22+(\d+6)*7.7;},
  head/.style={anchor=west, inner sep=0pt, font=\nzsize{8}{9.6}\bfseries},
  small/.style={font=\nzsize{6.5}{7.6}, inner sep=0pt},
  note/.style={small, text=nzink!62},
  tiny/.style={font=\nzsize{6}{7}, inner sep=0pt, text=nzink!62},
  axis/.style={draw=nzink, line width=0.5pt},
  tick/.style={draw=nzink, line width=0.4pt},
  ok/.style={->, >={Stealth[length=3pt]}, draw=nzink!78, line width=0.55pt},
  bad/.style={->, >={Stealth[length=2.6pt]}, draw=nzink!58, line width=0.5pt,
    dash pattern=on 1.7pt off 1.3pt},
  op/.style={fill=white, inner xsep=1.7pt, inner ysep=1pt, font=\nzsize{6}{7}},
  datafill/.style={fill=nzblue!16}, dataline/.style={draw=nzblue, line width=0.6pt},
  ghostfill/.style={fill=nzblue!4},
  ghostline/.style={draw=nzblue!80, line width=0.5pt, dash pattern=on 1.5pt off 1.1pt},
  chip/.style={rounded corners=1.5pt, line width=0.45pt, inner xsep=2.2pt, inner ysep=1.6pt,
    text height=4.2pt, text depth=1.2pt, font=\fontsize{6}{7}\selectfont\ttfamily},
  badge/.style={circle, draw=nzink!70, line width=0.45pt, inner sep=0pt, minimum size=7.8pt,
    font=\nzsize{6}{6}},
]
\path[use as bounding box] (0,4.4) rectangle (100,79.6);

\node[head] at (0,77.7) {(a) Dataset and Model Pixel Scales};

\foreach \o in {2.0,1.0}{
  \path[draw=nzblue!80, fill=white, line width=0.45pt] (1+\o,64.6+\o) rectangle ++(5.6,5.6);}
\foreach \i/\j/\t in {0/0/10,1/0/30,2/0/16,0/1/36,1/1/14,2/1/44,0/2/20,1/2/40,2/2/12}{
  \fill[nzblue!\t] ({1+\i*1.8667},{64.6+\j*1.8667}) rectangle ++(1.8667,1.8667);}
\draw[nzblue!80, line width=0.45pt] (1,64.6) rectangle ++(5.6,5.6);
\node[anchor=west, align=left, inner sep=0pt] at (10.4,68.4) {Dataset\\Delivers};

\path[draw=nzcoral!85, fill=nzcoral!8, line width=0.5pt, rounded corners=1.5pt]
  (1,45.4) rectangle ++(7.6,7.6);
\foreach \y in {47.3,49.2,51.1}{
  \path[draw=nzcoral!80, fill=nzcoral!30, line width=0.35pt, rounded corners=0.6pt]
    (2.3,{\y-0.55}) rectangle ++(5.0,1.1);}
\node[anchor=west, align=left, inner sep=0pt] at (10.4,49.2) {Model\\Expects};

\draw[ok] (24,69.6) -- (29.6,69.6);
\node[small, anchor=west] at (30.6,69.6) {Correct Conversion};
\draw[bad] (24,66.3) -- (29.6,66.3);
\node[small, anchor=west] at (30.6,66.3) {Wrong Conversion};

\nzbump{-1.6}{0}{63}{datafill}{dataline}
\nzbump{0.81}{2.41}{63}{datafill}{dataline}
\nzbump{2.4}{4}{63}{datafill}{dataline}
\draw[axis] ({X(-6)},63) -- ({X(4)},63);
\foreach \d in {-6,-4,-2,0,2,4}{\draw[tick] ({X(\d)},63) -- ++(0,-0.7);}
\node[small, anchor=base] at ({X(-0.8)},69.4) {Reflectance};
\node[small, anchor=base] at ({X(1.61)-1.3},69.4) {uint8};
\node[small, anchor=base] at ({X(3.2)+0.9},69.4) {S2 DN};
\node[tiny, anchor=base] at ({X(-0.8)},72.6) {m-so2sat};
\node[tiny, anchor=base] at ({X(1.61)-1.3},72.6) {m-pv4ger};
\node[tiny, anchor=base] at ({X(3.2)+0.9},72.6) {m-eurosat};

\path[fill=nzcoral!13, draw=nzcoral!75, line width=0.5pt]
  ({X(-1.75)},45) rectangle ({X(0.15)},51.5);
\nzbump{-1.6}{0}{45}{datafill}{dataline}
\nzbump{-5.6}{-4}{45}{ghostfill}{ghostline}
\nzbump{2.4}{4}{45}{ghostfill}{ghostline}
\draw[axis] ({X(-6)},45) -- ({X(4)},45);
\foreach \d/\l in {-6/\nzpow{\textminus 6},-4/\nzpow{\textminus 4},-2/\nzpow{\textminus 2},
    0/1,2/\nzpow{2},4/\nzpow{4}}{
  \draw[tick] ({X(\d)},45) -- ++(0,-0.7);
  \node[tiny, anchor=north, text=nzink] at ({X(\d)},43.7) {\l};}
\node[tiny, anchor=north east] at (17.4,43.7) {Pixel Value};
\node[tiny, anchor=west, align=left, text=nzcoral!72!black] at ({X(0.15)+1.2},48.1)
  {Pretraining\\Range};

\draw[bad] (57.8,61.9) -- node[op, pos=0.5] {\nzdiv\nzpow{4}} ({X(-4.8)},50.9);
\draw[ok] (60.2,61.9) -- node[op, pos=0.5] {\nzx\,1} (60.2,52.1);
\draw[ok] (76.4,61.9) -- node[op, pos=0.46] {\nzdiv 255} (63.6,52.1);
\draw[ok] (90.6,61.9) -- node[op, pos=0.27] {\nzdiv\nzpow{4}} (67.4,52.1);
\draw[bad] ({X(3.2)},61.9) -- node[op, pos=0.6] {\nzx\,1} ({X(3.2)},50.9);

\node[inner sep=0pt, anchor=north] (small) at ({X(-4.8)+1.4},40.3)
  {\nzpow{4}\nzx{} Too Small};
\node[inner sep=0pt, anchor=north] (match) at ({X(-0.8)+1.4},40.3) {Matched};
\node[inner sep=0pt, anchor=north east] (large) at (100,40.3) {\nzpow{4}\nzx{} Too Large};
\nzcross{([xshift=-0.022\linewidth]small.west)}
\nzcheck{([xshift=-0.020\linewidth]match.west)}
\nzcross{([xshift=-0.022\linewidth]large.west)}

\node[head] at (0,33.3) {(b) Choosing the Strategy for a New Model};

\node[badge] at (7.6,26.8) {1};
\path[draw=nzink!60, fill=white, line width=0.45pt, line join=round]
  (12.6,23.7) -- (12.6,29.9) -- (16.2,29.9) -- (17.6,28.5) -- (17.6,23.7) -- cycle;
\draw[nzink!60, line width=0.4pt, line join=round] (16.2,29.9) -- (16.2,28.5) -- (17.6,28.5);
\foreach \y in {27.4,26.2,25.0}{\draw[nzink!45, line width=0.4pt] (13.5,\y) -- (16.7,\y);}
\node[anchor=base] at (14,20.3) {Read the Release};
\node[note, anchor=base] at (14,16.9) {Paper, Code, Weights};

\node[badge] at (42.4,26.8) {2};
\foreach \i/\h/\c in {0/3.3/nzblue!70,1/2.3/nzblue!70,2/2.9/nzblue!70,3/5.6/nzcoral,4/1.7/nzink!35}{
  \fill[\c] ({47.3+\i*1.55},24.1) rectangle ++(1.05,\h);}
\draw[nzink!60, line width=0.4pt] (46.6,24.1) -- (55.2,24.1);
\node[anchor=base] at (50,20.3) {Test Candidates};
\node[note, anchor=base] at (50,16.9) {Evaluate Every Strategy};

\node[badge] at (77.6,26.8) {3};
\path[draw=nzink!60, fill=white, line width=0.45pt, line join=round]
  (82.4,24.2) rectangle ++(7.4,5.0);
\draw[nzink!60, line width=0.45pt, line join=round] (82.4,29.2) -- (86.1,26.4) -- (89.8,29.2);
\node[anchor=base] at (86,20.3) {Ask the Authors};
\node[note, anchor=base] at (86,16.9) {In a Public Issue};

\draw[ok] (23.4,26.8) -- node[tiny, above=1.2pt] {If Unclear} (36.6,26.8);
\draw[ok] (59.4,26.8) -- node[tiny, above=1.2pt] {To Confirm} (71.8,26.8);

\node[chip, anchor=west, draw=nzblue, fill=nzblue!8, line width=0.9pt] (c1) at (0.3,10.1) {dataset};
\node[chip, anchor=west, draw=nzblue!75, fill=nzblue!8] (c2)
  at ([xshift=0.016\linewidth]c1.east) {minmax};
\node[chip, anchor=west, draw=nzblue!75, fill=nzblue!8] (c3)
  at ([xshift=0.016\linewidth]c2.east) {minmax\_zscore};
\node[chip, draw=nzcoral!80, fill=nzcoral!9] (c4) at (68.4,10.1) {model};
\node[chip, draw=nzink!40, fill=nzsoft!25] (c5) at (90.6,10.1) {none};
\draw[nzink!45, line width=0.4pt] (50,15.4) -- (50,13.9);
\draw[nzink!45, line width=0.4pt, line join=round]
  (0.2,13.0) -- (0.2,13.9) -- (99.8,13.9) -- (99.8,13.0);
\draw[nzblue, line width=0.5pt] ([yshift=-0.012\linewidth]c1.west |- c2.south)
  -- node[small, below=1.6pt, text=nzink] {Dataset Statistics} ([yshift=-0.012\linewidth]c3.east |- c2.south);
\draw[nzcoral, line width=0.5pt] ([yshift=-0.012\linewidth]c4.west |- c2.south)
  -- node[small, below=1.6pt, text=nzink] {Pretraining Statistics} ([yshift=-0.012\linewidth]c4.east |- c2.south);
\draw[nzink!50, line width=0.5pt] ([yshift=-0.012\linewidth]c5.west |- c2.south)
  -- node[small, below=1.6pt, text=nzink] {No Rescaling} ([yshift=-0.012\linewidth]c5.east |- c2.south);
\end{tikzpicture}%
\endgroup

	\end{minipage}
    \caption{\textbf{Matching dataset and model pixel scales.} (a)~Datasets use different pixel scales, and models expect the scale used during pretraining. Dividing Sentinel-2 digital numbers (S2~DN) by $10^4$ produces the expected input range, but applying the same scaling to reflectance makes the values $10^4$ times too small. The curves are schematic, with offsets matching the stated conversion factors. (b)~We check each model's paper, code, and weights to identify its pretraining preprocessing. When unclear, we compare strategies on the validation split and contact the authors through GitHub issues to confirm our implementation.}
	\label{fig:normalization}
\end{supplementfigure}

In our harness, datasets return pixels as stored, without normalization, together with per-band statistics computed on the training split. The model wrapper handles preprocessing because the required units and normalization depend on the checkpoint. There are five options:
\begin{itemize}
    \item \texttt{dataset} (default): Normalize each band using its training-split mean and standard deviation. For reflectance stored as scaled integers, the scale factor cancels because it appears in the pixels, mean, and standard deviation. This gives standardized inputs, but uses the downstream dataset's statistics rather than those used during pretraining.
    \item \texttt{model}: Use the checkpoint's pretraining mean and standard deviation. Here, the pixel units must match the units of those statistics, so the wrapper converts the pixels first.
    \item \texttt{minmax} and \texttt{minmax\_zscore}: Both use the dataset's per-band ranges to rescale the pixels.
    \item \texttt{none}: Leave the pixels unchanged when normalization is already part of the model.
\end{itemize}

For SAR, the wrapper needs to know whether a band contains, for example, decibels or linear power before it can apply pretraining statistics. Datasets provide this information where available, and the wrapper performs the corresponding conversion. If the units are unknown, the wrapper either uses dataset statistics or rejects the band rather than assuming a unit.

When adding a model, we first check its paper, code, and released weights for the preprocessing used during pretraining. If these do not resolve the choice, we compare all five strategies on the validation split and retain the one with the highest score (\cref{fig:normalization}b).

\subsection{Solver comparison}
\label{subsec:solver-comparison}
We train the mini-batch Adam baseline with batches of 1,024 embeddings for at most 2,000 epochs and stop after the first epoch that does not improve the best epoch-averaged training objective by more than $10^{-6}$. The learning rate is constant within each fit, with no scheduler. Both solvers use eight regularization values, $C=10^k$ for $k=-4,\ldots,3$, and Adam also uses 12 learning rates, $\eta=a\,10^k$ for $a\in\{1,3\}$ and $k=-5,\ldots,0$. For each solver, we select the configuration with the highest validation accuracy. Full-batch L-BFGS uses a learning rate of 1, a strong Wolfe line search, and at most 2,000 iterations. Both solvers start from zero weights. For each fit, we copy the pre-extracted training embeddings and labels to the GPU once, and Adam draws its mini-batches there. Reported times include the transfer and classifier initialization but exclude feature extraction and prediction.

\section{Compute cost measurement}
\label{app:compute}

We measure FLOPs, parameter counts, throughput, latency, and memory with the harness's \code{flops} command under the setup below.

\paragraph{Counter.}
We count with PyTorch's operator-level counter, \code{torch.utils.flop\_counter.FlopCounterMode} (PyTorch 2.14.1), on one forward pass of the frozen model in evaluation mode, with a single float32 sample and no autocast. We count attention as its two batched matrix products, including where PyTorch would run it in a fused kernel.

\paragraph{Convention.}
The counter charges two operations per multiply--add. A product of an $m{\times}k$ and a $k{\times}n$ matrix costs $2mnk$, and a convolution costs $2\,C_\text{in}C_\text{out}\prod_i K_i\prod_j O_j$ for kernel size $K$, output size $O$, and $C_\text{in}$ input channels per group. Tools that report multiply--accumulates (MACs), such as fvcore, ptflops, and thop, give about half of our values for the same network, so our GFLOPs should be halved before they are compared with numbers from those tools. ResNet-50 at 224 pixels, for example, is 8.17 GFLOPs under our convention and about 4.1 under the MAC convention.

\paragraph{Coverage.}
The counter has formulas for matrix products, convolutions, and attention. Normalization layers, activations, pooling, interpolation, element-wise arithmetic, and embedding lookups contribute zero unless PyTorch decomposes them into one of these operators. The image-statistics and handcrafted baselines have no learned linear maps and report 0 GFLOPs, so we compare them by measured throughput.

\paragraph{Inputs.}
The inputs are synthetic Gaussian tensors of shape $1{\times}C{\times}S{\times}S$, with $C{=}3$ for RGB and $C{=}12$ for the Sentinel-2 bands. Models that use band metadata (sensor, wavelength, ground sampling distance) receive Sentinel-2 metadata, and SAR-only configurations receive Sentinel-1 metadata.

\paragraph{Reported quantities.}
A classification measurement covers the backbone and a linear probe, whose cost is negligible. A segmentation measurement covers the backbone, neck, and decoder at the same resolution, and its backbone share equals the classification backbone count for the same model and input. \Cref{tab:seg-lora} shows the segmentation total with the UPerNet decoder for an RGB input at 224 pixels.

\paragraph{Timing and memory.}
Throughput, median latency, and peak GPU memory come from the same forward pass at batch size 1, with five warm-up passes and twenty timed passes, each followed by a CUDA synchronize. They exclude data loading and host-to-device transfer. We took these measurements in float32 on one GPU model (NVIDIA RTX PRO 6000 Blackwell, 96\,GB).

\paragraph{Projected mapping time.}
We convert inference FLOPs into GPU-days assuming a sustained rate of 25.6\,TFLOP/s for every model. This is the median achieved rate of the 33 ranked models profiled at 512 pixels in float32 on one NVIDIA RTX PRO 6000 Blackwell. Actual throughput varies across architectures, so these estimates illustrate the scale of the computation rather than predict model-specific runtime. Note that these estimates are of the forward pass inference only and exclude data loading, tiling, and post-processing which are external costs.

\section{Dataset notes}
\label{app:datasets}

\Cref{tab:app-datasets-cls,tab:app-datasets-seg} give, per dataset, the splits, labels, and band layout we evaluate where they differ from the original release or from convention.

\begin{supplementtable}
	\centering
	\footnotesize
	\setlength{\tabcolsep}{3pt}
	\begin{tabularx}{\textwidth}{@{}L{3.3cm} Y@{}}
		\toprule
		Dataset                & Notes                                                                                                                                  \\
		\midrule
		\code{treesatai}       & Sentinel-1 is stored as whole-dB integers with a $-9999$ no-data marker (\cref{app:quirks}). \\
		\noterule
		\code{resisc45}        & Ground sampling distance (GSD) ranges from 0.2 to 30\,m, and models that take it as an input receive 1\,m because per sample GSD is unknown.                                  \\
		\bottomrule
	\end{tabularx}
	\caption{\textbf{Classification datasets.} Notes on the inputs as evaluated.}
	\label{tab:app-datasets-cls}
\end{supplementtable}

\begin{supplementtable}
	\centering
	\footnotesize
	\setlength{\tabcolsep}{3pt}
	\begin{tabularx}{\textwidth}{@{}L{3.3cm} Y@{}}
		\toprule
		Dataset                    & Notes                                                                                                                                                    \\
		\midrule
		\multicolumn{2}{@{}l}{\emph{GEO-Bench-2}}                                                                                                                                             \\
		\code{caffe}               & Single channel replicated to three, so every model can use its RGB input.                                                                                \\
		\noterule
		\code{fotw}                & Two-date image pairs upstream; we use the later image. Label 3 (unknown) is ignored.                                                                     \\
		\noterule
		\code{kuro\_siwo}          & Post-event SAR joined with the DEM. No RGB bands.                                                                                \\
		\noterule
		\code{spacenet7}           & One image per sample, so this is segmentation, not change detection.                                                                    \\
		\midrule
		\multicolumn{2}{@{}l}{\emph{Other}}                                                                                                                                                   \\
		\code{hotosm\_buildings}   & Trained on a fixed 5{,}000-tile subset of 57{,}890 (ranking agreement with full training: Kendall $\tau_b$ 0.97 frozen, 0.88 LoRA).                      \\
		\noterule
		\code{open\_cities}        & No labeled test set upstream, so we define chips and a 500\,m-block spatial split, and train on a fixed 10{,}000-chip subset of 76{,}818.                \\
		\noterule
		\code{openearthmap-sar}    & Label 0 is ignored, whereas the official baseline trains it as a ninth background class.                                                                 \\
		\noterule
		\code{levir-cd-post} & Post-change image only; excluded from the core benchmark. This is not a standard two-image change-detection evaluation. \\
		\noterule
		\code{sen1floods11-s1}     & Hand-labeled chips only. Both Sen1Floods11 variants ignore the union of each sensor's invalid pixels, so they score the same pixels.                     \\
		\noterule
		\code{bright-post-sar}     & Post-event SAR only.                                                                   \\
		\noterule
		\code{bright-post-sar-rgb} & The same tiles as three identical channels for RGB-only models.                                                                                          \\
		\bottomrule
	\end{tabularx}
	\caption{\textbf{Segmentation datasets.} Notes on splits, labels, and inputs as evaluated.}
	\label{tab:app-datasets-seg}
\end{supplementtable}

\section{Model notes}
\label{app:models}

\Cref{tab:app-models} lists the input handling that is specific to individual model families.

\begin{supplementtable}
	\centering
	\footnotesize
	\setlength{\tabcolsep}{3pt}
	\begin{tabularx}{\textwidth}{@{}L{4.3cm} Y@{}}
		\toprule
		Family                                 & Notes                                                                                                                                                                                                                                                                                              \\
		\midrule
		OlmoEarth v1 to v1.3                   & In classification, inputs keep each dataset's native size, upsampled to at least 64 pixels. v1.3 classification probes the location tokens of its attention pooler. Segmentation does not use the pooler, so we count v1.3 without it.                                                     \\
		\noterule
		Prithvi-v1, Prithvi-v2                 & Missing bands are dropped from the patch embedding.                                                                                                                                                                                                                                                \\
		\noterule
		Clay v1 base, v1.5 large               & Class-token and mean pooling give different $k$NN scores (\cref{app:quirks}).                                                                                                                                                                                  \\
		\noterule
		SMARTIES base, large                   & The Sentinel-2 processing level (L1C or L2A) used for pretraining normalization is set per dataset.                                                                                                                                                                          \\
		\noterule
		THOR tiny to large                     & Classification inputs keep each dataset's native size.                                                                                \\
		\noterule
		Panopticon                             & Input sides must be multiples of 14, so 256 and 512 become 252 and 518.                                                                                                                                                                                                                            \\
		\noterule
		SSL4EO MoCo                            & The 13-band MoCo ResNet-50 takes its bands by name, with B10 filled with zeros.                                                                                                                                                                       \\
		\noterule
		ScaleMAE large                         & Classification inputs are resized per dataset to between 32 and 288 pixels with area interpolation, with a matching per-dataset scale.                                                                                                                              \\
		\noterule
		UniverSat                              & Patch size is in meters.                                                                                                                                                                                                                                                                            \\
		\bottomrule
	\end{tabularx}
	\caption{\textbf{Model families.} Input handling specific to each family.}
	\label{tab:app-models}
\end{supplementtable}

\section{Findings from unified evaluation}
\label{app:quirks}

\paragraph{No-data values.}
In the Sentinel-1 rasters of TreeSatAI, the no-data marker of $-9999$ occurs in parts of only four samples. Including it in the band statistics would still raise the VV standard deviation from 3.6 to 105\,dB, which would make the standardized VV values about 29 times smaller. We replace marked pixels with the mean of the sample's valid pixels and leave the marker out of the statistics.

\paragraph{Input normalization.}
We evaluate each model with dataset standardization and, where it is defined, with model-native normalization (\cref{tab:choices}). With a linear probe, SatlasNet R-152 (SI) scores 59.4 on RESISC45 with dataset standardization and 80.4 with model-native normalization, and the 13-band MoCo ResNet-50 scores 79.1 and 97.6 on EuroSAT-Sp. We list the normalization that the main results use for each model in \cref{sec:supp-rules}.

\paragraph{Pooling.}
We evaluate both class-token and mean pooling for the two Clay models, the four Prithvi-v2 models, ScaleMAE large, the two SMARTIES models, and TerraFM-B. The main results use mean pooling for these models, and \cref{tab:choices} compares the two. The largest difference is for Clay v1 base on EuroSAT-Sp with all bands, where the class token scores 21.1 points higher than mean pooling with $k$NN.

\section{Result selection and ranking}
\label{sec:supp-rules}

\paragraph{Ranking.}
We rank the models on each dataset, give tied models the mean of their ranks, and average each model's ranks over the datasets. We rank a model only if it has a result on every dataset in the comparison. For RGB classification, both probes rank the same 100 models with complete results for both probes.

\paragraph{Normalization.}
The model comparisons in \cref{sec:analysis:settings} use one normalization per model. OlmoEarth, SatlasNet, SSL4EO MoCo, SeCo, GASSL, ScaleMAE, SMARTIES, TerraFM, and CROMA receive inputs normalized as in their pretraining. For every other model except the image-statistics baseline, we standardize each band with the dataset's training statistics. SMARTIES uses this standardization in segmentation on OpenEarthMap-SAR and SpaceNet2.

\paragraph{Segmentation.}
The runs behind \cref{tab:seg-lora} train for 30 epochs. We chose the LoRA learning rate for each dataset on its validation split, which gave $5\times10^{-4}$ on Burn Scars, CaFFe, CloudSEN12, FotW, Kuro Siwo, and the two Sen1Floods11 datasets, and $10^{-4}$ on the other seven core datasets. On each dataset a model receives the bands it accepts, so on multispectral datasets the general vision models use only the RGB bands while GFMs built for multispectral input use more.

\begin{supplementtable}
	\centering
	\footnotesize
	\setlength{\tabcolsep}{3pt}
\begin{tabular}{@{}llrrrr@{}}
\toprule
Choice & Comparison & Cells & Mean & Range of $\Delta$ & $\tau$ \\
\midrule
\multicolumn{6}{@{}l}{\emph{Classification}} \\
Probe & Linear $-$ $k$NN & 600 & +15.3 & [$-$4.3, +37.0] & 0.60 \\
Regularization & Swept $-$ $C{=}1$ & 25 & +1.6 & [$-$0.8, +7.8] & 0.67\rlap{$^a$} \\
Normalization & Native $-$ $z$-score & 1,120 & +1.8 & [$-$14.7, +37.5] & -- \\
Pooling, linear & CLS $-$ mean & 89 & $-$0.6 & [$-$7.3, +3.5] & -- \\
Pooling, $k$NN & CLS $-$ mean & 89 & +4.4 & [$-$1.6, +21.1] & -- \\
\midrule
\multicolumn{6}{@{}l}{\emph{Segmentation}} \\
Decoder & UPerNet $-$ UNet & 139 & +0.4 & [$-$1.5, +3.6] & -- \\
Head, 10 ep. & FPN $-$ linear & 124 & +3.9 & [$-$1.6, +47.8] & 0.40\rlap{$^b$} \\
Adaptation & LoRA $-$ frozen & 223 & +1.4 & [$-$1.4, +7.1] & 0.82\rlap{$^c$} \\
Adaptation & Full $-$ frozen & 223 & +1.5 & [$-$1.8, +7.7] & 0.69\rlap{$^c$} \\
Input size & 512 $-$ 224 px & 15 & +3.0 & [0.0, +7.0] & -- \\
Schedule & Variant $-$ 10 ep. & 140 & $-$0.3 & [$-$11.1, +5.3] & -- \\
\bottomrule
\end{tabular}

	\caption{ \textbf{Effect of individual evaluation choices.} Each row varies one setting while holding the others fixed. We report the mean and range of score differences ($\Delta \times 100$) across all (model, dataset) pairs, and Kendall's $\tau$~\cite{kendall1938tau} to measure agreement between model rankings. $^a$Five models across five GEO-Bench datasets. $^b$Lowest $\tau$ among pairs of four heads, evaluated on eight models. $^c$Ten models evaluated with frozen features, LoRA, and full fine-tuning.}
	\label{tab:choices}
\end{supplementtable}

\section{Effect of evaluation choices}
\label{sec:supp-choices}

\Cref{tab:choices} lists the comparisons that \cref{sec:choices} summarizes, and the paragraphs below describe the two studies behind its head and schedule rows.

\paragraph{Decoder heads.}
For the head comparison we train per-pixel linear, convolutional-block, FPN, and DPT heads for 10 epochs on frozen backbones, using eight models, eight GEO-Bench-2 datasets, and both RGB and all-band input. All four heads have a result in 124 (model, dataset, bands) cells.

\paragraph{Training schedules.}
The schedule comparison covers seven models on four datasets, with a 10-epoch cosine schedule at a learning rate of $10^{-3}$ as the reference. Four variants train for 30 epochs and one for 50, and three of the 30-epoch variants also lower the learning rate to $3\times10^{-4}$, raise it to $3\times10^{-3}$, or hold it constant. However, none of these variants raises the test mIoU by more than 0.1 points.

\par\medskip
\begin{nolinenumbers}
\noindent\begin{minipage}{\linewidth}
\section{Regularization grid}
\label{sec:supp-csweep}

\Cref{fig:csweep} shows the linear-probe test accuracy of five models on GEO-Bench ForestNet at each of the 41 values of $C$ compared in \cref{sec:choices}.

	\captionsetup{type=figure}
	\centering
	\includegraphics[width=0.48\linewidth]{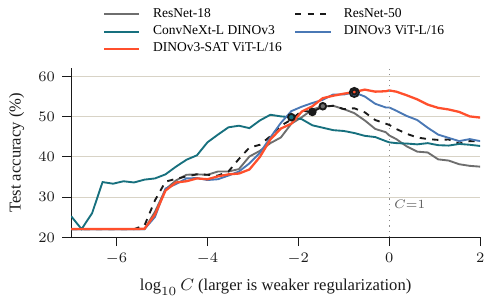}
	\caption{\textbf{Linear-probe accuracy across the regularization grid} on GEO-Bench ForestNet. Circles mark the $C$ chosen on the validation split. The dotted line marks $C=1$.}
	\label{fig:csweep}
\end{minipage}
\par\medskip

\noindent\begin{minipage}{\linewidth}
\section{Mapping compute by sensor group}
\label{sec:supp-sensor}

\Cref{fig:sensor-accuracy} shows, for each sensor group, the mean LoRA mIoU of seven models against the compute needed to map 1{,}000{,}000\,km$^2$, which \cref{sec:mapping} discusses.

	\captionsetup{type=figure}
	\centering
	\includegraphics[width=\textwidth]{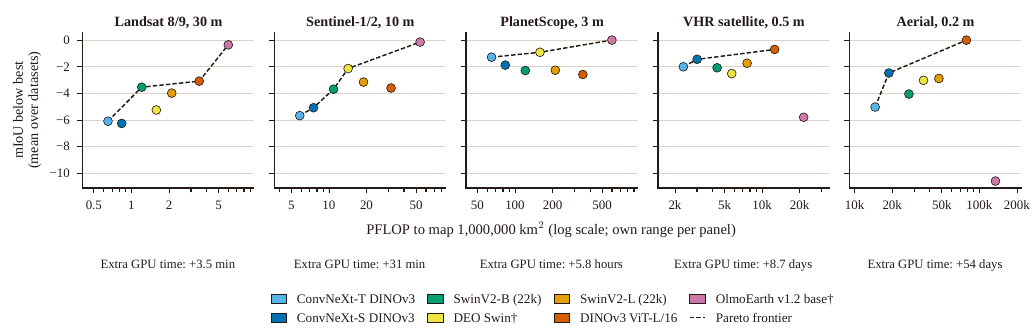}
    \caption{\textbf{LoRA accuracy versus compute required to map 1{,}000{,}000\,km$^2$, by sensor group.} Each panel uses a different compute range (0.6--6 PFLOP at 30\,m to 15k--134k PFLOP at 0.2\,m), since higher-resolution imagery requires more tiles. Accuracy is the mean LoRA mIoU across each group's core datasets, shown relative to the best of 34 ranked models. Three groups contain only one dataset. Compute assumes $224\times224$ RGB tiles at each group's nominal GSD, while accuracy uses the actual evaluation inputs, which may differ in size and bands. Labels below each panel show the GPU time saved by using the cheapest rather than the most expensive model, assuming 25.6\,TFLOP/s sustained throughput. $^\dagger$Pretrained on EO data.}
	\label{fig:sensor-accuracy}
\end{minipage}
\par\medskip
\end{nolinenumbers}

\end{document}